\documentclass[10pt,twocolumn,letterpaper]{article}

\usepackage{wacv}
\usepackage{times}
\usepackage{epsfig}
\usepackage{graphicx}
\usepackage{amsmath}
\usepackage{amssymb}

\usepackage{dsfont}
\usepackage{xfrac}
\usepackage{multicol}
\usepackage[table]{xcolor}
\usepackage{pifont}
\newcommand{\cmark}{\ding{51}}%
\newcommand{\xmark}{{\color{red} \ding{55}}}%

\usepackage{multirow}
\usepackage{tabularx}
\usepackage{booktabs}

\def\wacvPaperID{499} 

\wacvfinalcopy 

\ifwacvfinal
\def\assignedStartPage{9876} 
\fi

\ifwacvfinal
\usepackage[breaklinks=true,bookmarks=false]{hyperref}
\else
\usepackage[pagebackref=true,breaklinks=true,colorlinks,bookmarks=false]{hyperref}
\fi

\ifwacvfinal
\else
\fi

\newcommand*\samethanks[1][\value{footnote}]{\footnotemark[#1]}

\begin{document}

\title{Is An Image Worth Five Sentences? \\ 
A New Look into Semantics for Image-Text Matching}



\author{Ali Furkan Biten\thanks{Equal contribution.}  ~ ~ Andres Mafla\samethanks{} ~ ~ Lluis Gomez  ~ ~  Dimosthenis Karatzas\\
Computer Vision Center, UAB, Spain\\
{\tt\small {amafla, abiten, lgomez, dimos}@cvc.uab.es}}

\maketitle

\begin{abstract}
The task of image-text matching aims to map representations from different modalities into a common joint visual-textual embedding. However, the most widely used datasets for this task, MSCOCO and Flickr30K, are actually image captioning datasets that offer a very limited set of relationships between images and sentences in their ground-truth annotations.
This limited ground truth information forces us to use evaluation metrics based on binary relevance: given a sentence query we consider only one image as relevant. However, many other relevant images or captions may be present in the dataset.
In this work, we propose two metrics that evaluate the degree of semantic relevance of retrieved items, independently of their annotated binary relevance.
Additionally, we incorporate a novel strategy that uses an image captioning metric, CIDEr, to define a Semantic Adaptive Margin (SAM) to be optimized in a standard triplet loss. By incorporating our formulation to existing models, a \emph{large} improvement is obtained in scenarios where available training data is limited. We also demonstrate that the performance on the annotated image-caption pairs is maintained while improving on other non-annotated relevant items when employing the full training set. Code with our metrics and adaptive margin formulation will be made public.
\end{abstract}

\section{Introduction}

\begin{figure}
    \centering
    \includegraphics[width=\linewidth]{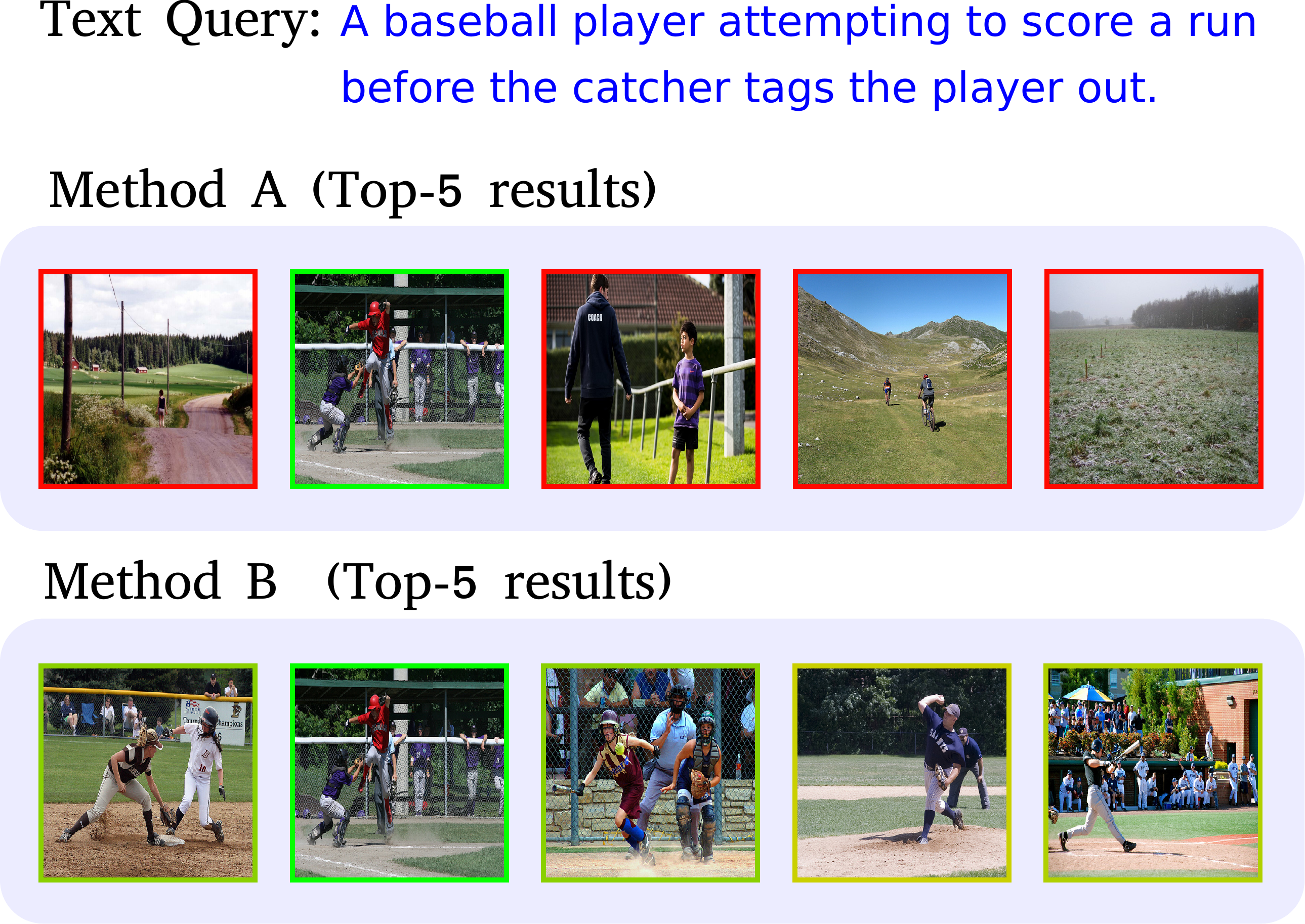}
    \caption{According to the Recall@5 metric, defined for Image Text Matching, both methods A and B are equally good: it considers only one image as relevant for a given sentence query. In this paper we present two metrics and an adaptive margin loss that takes into account that there might be other relevant images in the dataset. In this Figure we represent the semantic similarity of images to the query by their colored border (the greener the more similar).}
    \label{fig:my_label}
\end{figure}

Language provides a medium to explain our perceptual experience while being discretely infinite. ``Discrete infinity'' is referred as a property in which language is constructed by using few discrete elements albeit giving infinite variety of interpretations~\cite{chomsky2006architecture,studdert2003launching}. 
In other words, the language's discrete infinity property dictates that a potentially infinite number of semantically correct sentences can be used to express the same idea, for example, at describing an image. 
Framing the previous notion into consideration, we explore the task of Image-Text Matching (ITM) in a cross-modal retrieval scenario. Image-text matching refers to the problem of retrieving a ranked list of relevant representations of the query portrayed by a different modality. Yet, somehow contrary to the notion of discrete infinity, commonly used datasets for the image-text matching (ITM) task lack exhaustive annotations of many-to-many mappings between images and captions. These datasets are designed originally for the image captioning task. Nonetheless, the assumption in ITM is that only 5 sentences correctly describe a single image, labeling it in a binary manner as relevant or irrelevant.
Consequently, the lack of many-to-many annotations causes a direct effect on the way the ITM task is evaluated. Sentences that are not relevant according to the ground-truth can describe an image in various degrees of correctness and coverage, thus making the way we evaluate current models on ITM \emph{incomplete}. We can see an example of such problem in Figure~\ref{fig:my_label}. The widely adopted metric employed to evaluate the performance of a model in the ITM task is Recall@K~\cite{faghri2017vse++, lee2018stacked, li2019visual, li2020oscar, liu2020graph, lu202012}. The Recall@K as it is used in IMT is binary by definition: it returns $1$ if at least one of the relevant items according to the ground truth is retrieved within the top-k positions for a given query, otherwise it returns $0$. Due to this binary definition, the metric can not fully assess the degree of accuracy and coverage of the retrieved sentences given an image or the other way around.

\begin{figure*}
    \centering
    \includegraphics[width=\textwidth]{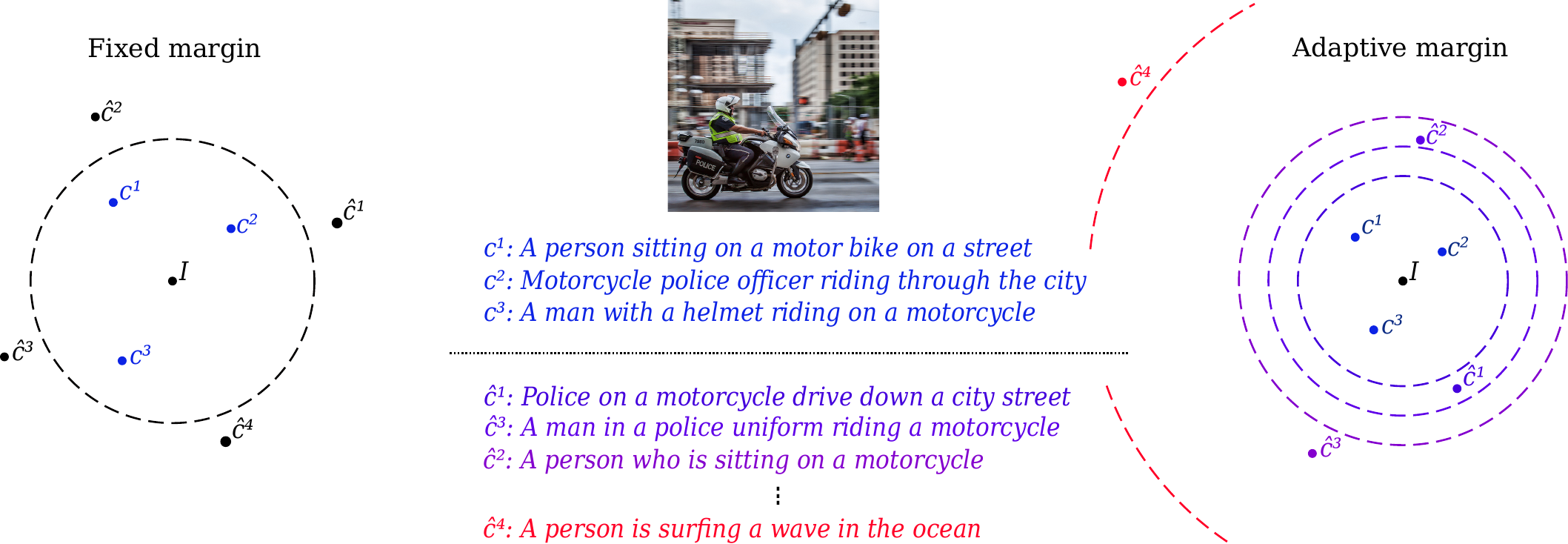}
    \caption{Comparison of a fixed margin loss function (left) and our adaptive margin (right). We consider an image anchor $I$, their positive sentences according to the ground truth ($c^1, c^2, c^3$), and four other sentences ($\hat{c}^1, \hat{c}^2, \hat{c}^3, \hat{c}^4$) that are negative according to the ground-truth but have some degree of semantic similarity with $I$. In our method we dynamically adapt the margin of each possible triplet (anchor, positive, and negative items) to the value given by a similarity function $\phi$ that measures the semantic similarity of positive and negative items. In this Figure we represent the similarity of sentences with the anchor by its color (the bluish the more semantically similar they are, the reddish the less similar).}
    \label{fig:adaptive_margin}
\end{figure*}

Additionally, the to-go strategy from standard approaches for ITM, firstly introduced by~\cite{faghri2017vse++}, relies on hard-negative mining at the moment of constructing samples to be used in a Triplet loss function to be optimized. Current state-of-the art methods rely heavily on such formulation, which requires a carefully handcrafted fixed margin~\cite{faghri2017vse++, lee2018stacked, li2019visual, liu2020graph, mafla2021stacmr, wang2019camp}. 
In this work, we propose solutions to the aforementioned problems by introducing the usage of image captioning metrics such as SPICE~\cite{anderson2016spice} and CIDEr~\cite{vedantam2015cider} as a part of an additional metric formulation into the evaluation of the ITM task. Image captioning metrics have been widely studied and accepted as automatic tools to evaluate the similarity of sentence meanings that closely correlate with human judgement. We utilize such metrics 
that allows a transition from a traditional recall to a Normalized Cumulative Semantic (NCS) Recall by incorporating the continuum of language into the evaluation. Secondly, considering 
the continuous nature of language, we re-formulate a triplet loss by introducing a Semantic Adaptive Margin (SAM). We calculate a SAM according to image captioning metrics, which does not rely on a hard-negative mining approach (see Figure~\ref{fig:adaptive_margin}). Our formulation employed in scenarios with limited data achieves state of the art by a \emph{significant} retrieval improvement.



Our contributions are as follows: (1) We identify shortcomings from the commonly used Recall$@K$ in the ITM task. By adopting image captioning metrics we model the many-to-many semantic relationships between images and captions. (2) We propose a novel Semantic Adaptive Margin (SAM) that takes into consideration image captioning metrics to define the similarity among samples. (3) We show that by relying on image captioning metrics and incorporating them into our proposed adaptive margin, a substantial boost is achieved in scenarios with reduced training data. (4) We provide exhaustive experiments on two benchmark datasets, which show that by incorporating our adaptive margin formulation an increase in performance is achieved across a variety of state-of-the-art pipelines. 





\section{Related Work}
\label{sec:related_work}
\textbf{Cross-modal retrieval.}
Our proposed work focuses on the task of cross-modal retrieval, particularly on image-text matching. The task aims to map the images and sentences in such a way that a suitable space for retrieval is learned, where the query and the search data come from distinct modalities.


Initial approaches~\cite{faghri2017vse++, frome2013devise} learned to align the global visual and textual features 
by applying a learned non-linear transformation to project both modalities into a common space. A similar pipeline is proposed by~\cite{nam2017dual} with the incorporation of an attention mechanism. 
However, the main drawback of such approaches 
is that semantic concepts fall short at capturing fine-grained semantic interactions between visual regions and sentences. 
In the work presented by~\cite{anderson2018bottom}, several visual regions that describe an image in a more detailed manner are used for the task of Visual Question Answering (VQA) and Image Captioning. 
Initial works~\cite{niu2017hierarchical} incorporated visual regions along with a hierarchical Long Short-Term Memory (LSTM)~\cite{hochreiter1997long} module. 
Following up,~\cite{lee2018stacked} proposed a stacked cross attention network to model the latent alignments between image regions and words.
Additional models have explored the roll of attention mechanisms~\cite{liu2019focus, song2019polysemous, wang2019camp, wei2020multi, zhang2020context},  
and Graph Convolutional Neural Networks (GCN)~\cite{diao2021similarity, kipf2016semi, li2019visual, liu2020graph}.
External modules have been explored to improve retrieval results such as the usage of an iterative recurrent attention module~\cite{chen2020imram} and an external consensus knowledge base~\cite{wang2020consensus}. 
More recently, Transformers~\cite{vaswani2017attention} have been used to learn intra and inter-modality attention maps for a wide range of visual and language tasks~\cite{li2020oscar, jia2021scaling, lu202012, lu2019vilbert}, often achieving state of the art. However, these approaches require an additional order of magnitude of training samples, giving rise to a large increase in computational costs. In our work, we focus on task-specific architectures that directly employ the joint embedding space at retrieval time and surpasses by a huge margin current state-of-the-art approaches while using \textit{significantly} less training samples.

\textbf{Semantics and Metric Learning.}
Closely related to our work,~\cite{musgrave2020metric} highlights the main flaws of current metric learning approaches, in which shows that
metrics are not consistent for the task at hand. Also it is shown that the gap between methods is less significant when evaluation is properly done. In this work we refer to the problem of captions that can correctly describe an image that is not annotated in the GT, as semantic gap.
Trying to overcome the existing semantic gap in current datasets,~\cite{huang2018learning} employs a network to predict the semantic concepts from an image, however they rely to a binary annotation of relevance. Other works~\cite{gomez2017self, gordo2017beyond} propose a model to learn a visual embedding where the image similarity is correlated with paired text.
Similarly~\cite{thomas2020preserving} proposes a novel within-modality loss that leverages semantic coherency between text and images, which do not necessarily align with visually similar images. In order to address non-paired relevant images and captions,~\cite{zhang2020learning} proposes to build denotation graphs to link between these two modalities. Trying to overcome the non-exhaustive annotation in datasets, ~\cite{chun2021probabilistic} models the probability of an image belonging to a set of specific contexts. A newly introduced CrissCrossed~\cite{parekh2020crisscrossed} dataset, is an extension of MS-COCO that collects human judgements on the degree of how an image and a sentence matches. 
Closely related to our approach,~\cite{zhou2020ladder} proposes the usage of a ladder loss among samples based on BERT~\cite{devlin2018bert} to define similarities. However, calculating BERT for every sample is very expensive computationally, thus they rely on a threshold given by a CBOW~\cite{wang2018glue} to refine the comparison. Another drawback is that the similarity is computed among two captions alone, thus not all information is leveraged. 

In our work, we present several metrics that show the limitations of current cross-modal retrieval methods. We also solve previously defined limitations by employing $5$ captions to compute a richer semantic similarity while still being a very cheap computationally alternative. Consequently, we incorporate a Semantic Adaptive Margin (SAM) based on automatic image captioning metrics. 

\textbf{Image Captioning evaluation metrics.}
Image captioning is the task of transcribing an image into natural language. 
There are metrics proposed specifically for image captioning models, specifically, CIDEr~\cite{vedantam2015cider} and SPICE~\cite{anderson2016spice}. These recently proposed metrics not only have been widely accepted to evaluate the captioning models, but also they have been shown to correlate better with human judgments across different datasets~\cite{anderson2016spice, vedantam2015cider} when compared to machine translation metrics. 
CIDEr proposes tf-idf~\cite{jones1972statistical} to weight the importance of each n-gram. Moreover, CIDEr employs cosine distance between tf-idf values of reference and hypothesis instead of using direct match, which accounts for both precision and recall. SPICE is the first captioning metric that does not follow the classic n-gram matching. The SPICE score is an F-score calculated by direct matching between the semantic scene graphs of reference and hypothesis. 
Despite their strengths, CIDEr and SPICE still have certain limitations regarding synonyms, word order, style and sentence grammar among others ~\cite{elliott2014comparing, Kilickaya2016}. Aside from the limitations, they remain as good automatic metrics to measure semantic similarity, specially when data comes from a similar distribution. Hence we employ the aforementioned metrics into a classical retrieval scenario.
In the following section, we will describe our motivation and the need of new insights by proposing a new metric and explain each of the variants.





\section{Metrics}
\label{sec:Metrics}
In this section, we present our proposed metrics which aim to offer additional insights on the performance amongst models designed to tackle the ITM task. 
Before we move with our formulation, we introduce the reader the nomenclature used in the rest of this work. 
First, the image and caption sets will be referred to as $I$ and $C$, while the respective test set will be represented by $I_T$, $C_T$. 
We refer as $G_{l}$ to all the ground truth captions corresponding to an image $l \in I$. We use $\phi$ to indicate an evaluation metric function such as CIDEr or SPICE. Finally, $Q_{ik}$ represents the \textit{retrieved} items for a given query $i$ at a top-$k$ cut-off threshold.


\subsection{Is an image worth 5 sentences?}
\label{subsec:image5sentences}
Both of the most commonly used datasets in ITM, namely Flickr30k~\cite{plummer2015flickr30k} and MSCOCO~\cite{lin2014microsoft}, contain 5 ground truth sentences per image. A direct outcome is that current evaluation solely considers those 5 sentences as relevant to a single image for the ITM task. However, it is a known fact that in MSCOCO or Flickr30k there are many sentences that can perfectly describe a non paired image~\cite{parekh2020crisscrossed, wang2020compare, wang2020towards}. In other words, there are sentences (images) that are relevant to images (sentences) even though they are not defined as such in the retrieval ground truth. We refer to these samples as non ground truth (non-GT) relevant items.  
Specifically, ITM models are tested on 5k images and 25k sentences in MSCOCO. In the case of image-to-text retrieval, recall completely ignores the retrieved order of the remaining 24995 sentences (99.98\% of the test set). Nevertheless, it is crucial to consider all semantically relevant items (including non-GT) to properly evaluate a models' capability.

Aside from the aforementioned problems, Recall@K (R@K) as used in the ITM task is a binary metric, i.e. it is a hard metric that does not take into account the semantic continuum of language. When it comes to language, even ground truth paired sentences do not explain a given image in the same degree as they are not paraphrases from each other. 

Another identified drawback is that the recall formulation used in ITM is different than the original recall employed in information retrieval. The recall metric used in ITM, referred as $R^V$, takes the definition from~\cite{hodosh2013framing}. In the image-to-text scenario, the $R^V$ only cares about the first GT annotated caption retrieved in the top-$k$ relevant results. This formulation discards the remaining $4$ annotated samples in the GT. On the other hand, recall defined by~\cite{schutze2008introduction}, referred as $R$, considers \textit{all} other relevant items in the formulation. It is important to note that both formulations agree on the text-to-image scenario due to the existence of only $1$ relevant image in the GT. Both recall formulations can be appreciated better in Equation~\ref{eq:vse_recall} and~\ref{eq:recall}.

\begin{equation}
  R@k = \frac{1}{|I_T|} \sum_{i\in I_T} R_i@k, \ \text{where} \ R_i@k = \frac{|G_i  \cap Q_{ik}|}{|G_i|}
\label{eq:vse_recall}
\end{equation}

\begin{equation}
    \ R^V@k = \frac{1}{|I_T|} \sum_{i\in I_T} R_i^V@k, \ \text{where} \ R_i^V@k = 
    \mathds{1}_{\{G_i  \cap Q_{ik} \ne \emptyset\}}
\label{eq:recall}
\end{equation}

When formulating our metrics in the following sections, we use $R$ instead of $R^V$, as it includes the remaining $4$ items at evaluation. Nevertheless, it is important to note here that both $R$ and $R^V$ completely disregard the possible semantic relevance of non-GT samples. The existent limitations of employing solely recall as a metric lie in the fact that it misses to evaluate those non-GT relevant items.

\subsection{Semantic Recall (SR)}
\label{subsec:SR}

Our metrics rely on the evaluations of captions with CIDEr and SPICE (for more details refer to Section~\ref{sec:related_work}) to decide which images are semantically similar to other sentences in the test set. Concretely, for a given image $i$ and sentence $j$ such that $i \in I_T, j \in C_T$, we construct a matrix $N$ where: 

\begin{equation}
N_{ij} = \phi(G_i, c_k)
\label{eq:metric}   
\end{equation}

\noindent where $\phi$ is one of the aforementioned captioning metrics (CIDEr or SPICE). Once the similarity matrix $N$ is defined, we can easily extend the ground truth relevant items for each possible query. Formally, we define $\Tilde{G_i}$ as the extension of ground truth relevant items for a query image $i$ as the most similar $m$ sentences from $N_i$. Now we define the Semantic Recall (SR) metric as follows:

\begin{equation}
      R_i^{SR}@k = \frac{|\Tilde{G_i}  \cap Q_{ik}|}{|\Tilde{G_i}|}
\end{equation}

This metric allows a transition from the classic recall $R^{V}$ to a metric that considers semantic relevance. However, the limitation on binary scoring associated with recall still persists. Another drawback is how to select a threshold $m$ that captures how many non-GT images or sentences are relevant in the whole data corpus. 

\subsection{Normalized Cumulative Semantic (NCS) Score}
\label{subsec:NCS}
The Normalized Cumulative Semantic Score (NCS) aims at addressing the limitations of the Semantic Recall (SR) described in previous section. The NCS score is calculated as the division between the image captioning similarity $\phi$ of the retrieved samples and the maximum image captioning similarity score $\phi$ at a cut-off point K. Formally, we define our metric as:

\begin{equation}
      N_{i}@k =\frac{\sum_j N_{ij}}{\sum_l N_{il}}, 
      \text{for} \ j \in \Tilde{G_i} \cap Q_{ik}\ \text{and}\ l \in \Tilde{G_i}
\end{equation}

For illustrative purposes, method A and B from Figure~\ref{fig:my_label}, both equally good at recall ($R^V$), will score very differently at NCS. Method A will achieve a maximum score of $0.2$. On the contrary, Method B will achieve a higher score since the retrieved samples contain a closer degree of semantics compared to the query.

With this formulation, we specify a solution to the binary nature of Recall@K (R@K) when it addresses the semantics of language. Moreover, NCS can properly take into account the non-GT items when evaluating a model without the need of selecting a threshold $m$. In Section~\ref{sec:Experiments} we use these metrics to provide us with additional insights about current model's performance.


\subsection{Correlation with Human Judgements}
\label{Correlation_CrissCrossed}
Related to our work, the recently introduced CrissCrossed~\cite{parekh2020crisscrossed} dataset, is an extension of MS-COCO that comprises human judgements on the degree of similarity between captions and images.
In this dataset, each annotator assesses how well an image and a sentence match on a 5-point likert scale on MSCOCO. They collect these judgements not only for the predefined ground truth pairs but also for other pairs. Despite the \textit{extensive} annotation process required, the test set contains 44k judgement pairs, of which 25k are ground truth pairs. 

We utilize these human judgments to calculate Pearson-R correlation coefficient for Recall and NCS. 
As it can be seen in~\autoref{table:correlation}, when all the pairs are considered, our metric has a better correlation with human judgments~\cite{anderson2016spice} with both SPICE and CIDEr. We observe that CIDEr has a better correlation when we take into account the 44k pairs, nonetheless SPICE is better on Non-GT, Which is why we always evaluate our models with SPICE.  Furthermore, this also extends to the case of non-ground truth relevant pairs. In non-GT relevant pairs the classic recall is uninformative due to the metric definition, while the NCS provides an acceptable estimation that correlates well with human judgement.


\begin{table}[htp]
\centering
\footnotesize
\begin{tabularx}{0.8\linewidth} { X c c }
   \toprule
    & All & Non-GT \\
   \midrule
   \textbf{Binary relevance (GT)} & 0.711 & 0.00 \\
   \textbf{NCS with SPICE} & 0.729 & \textbf{0.536} \\
   \textbf{NCS with CIDEr} & \textbf{0.734} & 0.453 \\
   \bottomrule
\end{tabularx}
\caption{Pearson-R correlation coefficient results between human judgements and image text matching metrics on the CrissCrossed~\cite{parekh2020crisscrossed} dataset.}
\label{table:correlation}
\end{table}


\section{Methodology}
In this section we introduce our Semantic Adaptive Margin (SAM) formulation, which aims to alleviate common problems of the usage of a triplet loss on non-exhaustive many-to-many data mappings.
Before we elaborate on the details, we present the reader with the original triplet formulation along with a formal definition of the ITM task.

Let $D= \{(i_n, c_n)\}_{n=1}^{N}$ be the training set of image and caption pairs. These pairs are further divided into positive and negative samples where $(i_p, c_p)$ are considered as positive samples while $(i_k, c_m)_{(k,m)\ne p}$ as negative samples. 
Then, the embedded images and captions are represented as $e_{c_p} = \sigma_c(c_p)$ and $e_{i_p} = \sigma_i(i_p)$ where $\sigma_c, \sigma_i$ are embedding functions for captions and images respectively. Given 
a similarity function $\psi$, the classic formulation of the triplet loss in ITM~\cite{faghri2017vse++}, $L_T$, is defined as:
\begin{equation}
\begin{aligned}[b]
    L_T = &max[\alpha + \psi(e_{i_p}, e_{c_m}) - \psi(e_{i_p}, e_{c_p}), 0] \
    + \\ 
    &max[\alpha + \psi(e_{i_k}, e_{c_p}) - \psi(e_{i_p}, e_{c_p}), 0] 
\end{aligned}
\label{eq:triplet}
\end{equation}

\noindent
where $\alpha$ is known as the margin. The intuition behind the triplet formulation is that given an n-sphere with radius $\alpha$, positive samples should be projected inside and negative samples on the external region of the n-sphere. This can be observed in the left section of~\autoref{fig:adaptive_margin}. It is important to remark that the margin employed in the triplet loss is fixed despite the relatedness of hard-negative pairs.

\subsection{Semantic Adaptive Margin (SAM)}
Even though a fixed margin might be acceptable in image-to-image metric learning tasks, a fixed margin can not capture the continuum of language properly. Looking at the right on~\autoref{fig:adaptive_margin}, we can acknowledge that even the non-GT items can properly explain the provided image. Therefore, using a fixed margin and treating every negative as equal is unfeasible if the semantics is to be modelled properly. Due to this fact, creating an adaptive margin is imperative to teach our models the continuous nature of language. 

Consequently, we formulate the Semantic Adaptive Margin (SAM) to dynamically calculate the similarity between images and sentences. More formally, given a positive pair $i_p, c_p$ with negative samples $(i_k, c_k), (i_m, c_m)$, we use the ground truth caption set $G_p$ to calculate the triplet loss by incorporating a SAM ($L_{SAM}$):
\begin{equation}
\begin{split}
    L_{SAM} &= max[\alpha_{i2t} + \psi(e_{i_p}, e_{c_m}) - \psi(e_{i_p}, e_{c_p}), 0] \ + \\ & \ \ \ \ \ max[\alpha_{t2i} + \psi(e_{i_k}, e_{c_p}) - \psi(e_{i_p}, e_{c_p}), 0] \\
    \alpha_{i2t} &= (\phi(G_p, c_p) - \phi(G_p, c_m))/\tau \\ 
    \alpha_{t2i} &=  (\phi(G_p, c_p) - \phi(G_p, c_k))/\tau \\ 
\end{split}
\label{eq:Adap_margin}
\end{equation}

\noindent
where $\psi$ is a similarity function such as cosine similarity, $\phi$ stands for an aforementioned captioning metric (SPICE or CIDEr) and $\tau$ is a temperature parameter to be controlled on how wide or small the margin is desired. In other words, $\tau$ is used as a scaling factor. 
As it can be appreciated by Equation~\ref{eq:Adap_margin}, we incorporate a SAM into the original triplet formulation, which assigns a unique margin value specific to each sampled pair. SAM still can be optimized jointly with the original triplet formulation.


\section{Experiments}
\label{sec:Experiments}
\definecolor{gray}{RGB}{210,210,210}
\newcolumntype{a}{>{\columncolor{gray}}c}

\begin{table*}[!h]
\begin{center}
\scriptsize

\begin{tabular}{|c|l|rrr|rrr|r|rrr|rrr|r|}
\hline
\multicolumn{2}{|c|}{\multirow{3}{*}{\textbf{Method}}} & \multicolumn{7}{c|}{\textbf{Recall}} & \multicolumn{7}{c|}{\textbf{Normalized Cumulative Semantic Score}} \\ \cline{3-16} 
\multicolumn{2}{|c|}{} & \multicolumn{3}{c|}{\textbf{I2T}} & \multicolumn{3}{c|}{\textbf{T2I}} & \multicolumn{1}{l|}{} & \multicolumn{3}{c|}{\textbf{I2T}} & \multicolumn{3}{c|}{\textbf{T2I}} & \multicolumn{1}{l|}{} \\ \cline{3-16} 
\multicolumn{2}{|c|}{} & \multicolumn{1}{c}{\textbf{R@1}} & \multicolumn{1}{c}{\textbf{R@5}} & \multicolumn{1}{c|}{\textbf{R@10}} & \multicolumn{1}{c}{\textbf{R@1}} & \multicolumn{1}{c}{\textbf{R@5}} & \multicolumn{1}{c|}{\textbf{R@10}} & \multicolumn{1}{c|}{\textbf{Rsum}} & \multicolumn{1}{c}{\textbf{N@1}} & \multicolumn{1}{c}{\textbf{N@5}} & \multicolumn{1}{c|}{\textbf{N@10}} & \multicolumn{1}{c}{\textbf{N@1}} & \multicolumn{1}{c}{\textbf{N@5}} & \multicolumn{1}{c|}{\textbf{N@10}} & \multicolumn{1}{l|}{\textbf{Nsum}} \\ \hline
\multirow{4}{*}{\rotatebox[origin=c]{90}{\textbf{F-10\%}}} & CVSE~\cite{wang2020consensus}$\dagger$ & 13.0 & 16.2 & 23.9 & 12.5 & 30.9 & 42.2 & 138.7 & 19.0 & 24.4 & 28.3 & 25.7 & 34.0 & 36.3 & 167.6 \\
 & CVSE+\scriptsize{SAM} & \textbf{34.6} & \textbf{60.4} & \textbf{70.9} & \textbf{23.9} & \textbf{50.6} & \textbf{62.8} & \textbf{303.2} & \textbf{37.4} & \textbf{40.2} & \textbf{44.1} & \textbf{38.6} & \textbf{46.3} & \textbf{47.1} & \textbf{253.6} \\ \cline{2-16} 
 & SGR~\cite{diao2021similarity}$\dagger$ & 0.3 & 0.7 & 1.3 & 0.2 & 0.6 & 1.1 & 4.2 & 2.8 & 4.2 & 5.7 & 5.7 & 9.7 & 11.7 & 39.8 \\
 & SGR+\scriptsize{SAM} & \textbf{37.9} & \textbf{64.8} & \textbf{77.5} & \textbf{26.6} & \textbf{53.4} & \textbf{64.4} & \textbf{324.6} & \textbf{40.1} & \textbf{41.0} & \textbf{44.6} & \textbf{41.4} & \textbf{48.1} & \textbf{48.5} & \textbf{263.7} \\ \hline
\multirow{4}{*}{\rotatebox[origin=c]{90}{\textbf{F-25\%}}} & CVSE~\cite{wang2020consensus}$\dagger$ & 30.4 & 47.5 & 59.6 & 28.8 & 58.1 & 69.7 & 294.1 & 30.8 & 39.9 & 43.7 & 42.4 & 48.0 & 47.7 & 252.5 \\
 & CVSE+\scriptsize{SAM} & \textbf{48.7} & \textbf{73.2} & \textbf{81.8} & \textbf{37.9} & \textbf{66.2} & \textbf{76.0} & \textbf{383.8} & \textbf{46.2} & \textbf{49.7} & \textbf{52.4} & \textbf{51.0} & \textbf{52.7} & \textbf{50.9} & \textbf{302.8} \\ \cline{2-16} 
 & SGR~\cite{diao2021similarity}$\dagger$ & 11.0 & 29.3 & 40.0 & 7.3 & 21.4 & 31.6 & 140.6 & 20.3 & 24.1 & 29.7 & 22.7 & 33.4 & 37.3 & 167.6 \\
 & SGR+\scriptsize{SAM} & \textbf{54.0} & \textbf{81.4} & \textbf{87.8} & \textbf{41.3} & \textbf{68.0} & \textbf{77.5} & \textbf{410.0} & \textbf{52.7} & \textbf{52.5} & \textbf{53.8} & \textbf{54.2} & \textbf{53.9} & \textbf{51.9} & \textbf{319.0} \\ \hline
\multirow{4}{*}{\rotatebox[origin=c]{90}{\textbf{C-3\%}}} & CVSE~\cite{wang2020consensus}$\dagger$ & 28.2 & 49.0 & 63.6 & 24.9 & 58.5 & 74.4 & 298.6 & 34.9 & 40.6 & 43.3 & 44.6 & 57.5 & 59.8 & 280.9 \\
 & CVSE+\scriptsize{SAM} & \textbf{41.1} & \textbf{71.8} & \textbf{82.1} & \textbf{32.2} & \textbf{65.4} & \textbf{77.2} & \textbf{369.8} & \textbf{47.2} & \textbf{49.5} & \textbf{51.6} & \textbf{50.6} & \textbf{59.8} & \textbf{59.8} & \textbf{318.5} \\ \cline{2-16} 
 & SGR~\cite{diao2021similarity}$\dagger$ & 0.2 & 1.1 & 2.1 & 0.1 & 0.6 & 1.2 & 5.3 & 4.1 & 5.1 & 5.9 & 4.2 & 6.7 & 8.1 & 34.2\\
  & SGR+\scriptsize{SAM} & \textbf{23.7} & \textbf{59.0} & \textbf{74.3} & \textbf{24.9} & \textbf{56.4} & \textbf{72.3} & \textbf{310.6} & \textbf{33.8} & \textbf{39.2} & \textbf{43.5} & \textbf{45.3} & \textbf{58.6} & \textbf{61.7} & \textbf{282.1} \\ \hline
\multirow{4}{*}{\rotatebox[origin=c]{90}{\textbf{C-5\%}}} & CVSE~\cite{wang2020consensus}$\dagger$ & 48.5 & \textbf{77.8} & 85.7 & 36.2 & 69.7 & 81.6 & 399.5 & \textbf{51.4} & 53.8 & 54.8 & 54.5 & 62.7 & \textbf{62.3} & 339.7\\
 & CVSE+\scriptsize{SAM} & \textbf{48.6} & 77.3 & \textbf{86.5} & \textbf{37.9} & \textbf{71.1} & \textbf{82.6} & \textbf{404.0} & 50.7 & \textbf{54.2} & \textbf{55.4} & \textbf{55.9} & \textbf{62.9} & \textbf{62.3} & \textbf{341.5} \\ \cline{2-16} 
 & SGR~\cite{diao2021similarity}$\dagger$ & 1.0 & 3.0 & 5.3 & 0.2 & 0.5 & 1.3 & 11.3 & 6.9 & 7.9 & 9.0 & 2.9 & 4.2 & 5.7 & 36.6 \\
 & SGR+\scriptsize{SAM} & \textbf{30.4} & \textbf{62.6} & \textbf{79.1} & \textbf{29.4} & \textbf{63.1} & \textbf{77.6} & \textbf{342.2} & \textbf{39.2} & \textbf{42.6} & \textbf{46.9} & \textbf{49.6} & \textbf{61.3} & \textbf{63.6} & \textbf{303.3} \\ \hline
 
\end{tabular}

\end{center}
\caption{Quantitative results on reduced training data samples. The acronyms used in the first column stand for Flickr30K (\textit{F}), MSCOCO 1K (\textit{C}). The (\%) denotes the proportion of the training data used in relation to the original dataset size. Results are depicted in terms of Recall@K (R@K) and Normalized Cumulative Semantic Score (N@K). The $\dagger$ depicts that models are trained with the publicly available code released by the original authors.}

\label{tab:minidata_results}
\end{table*}

In this section we present the results obtained by evaluating state-of-the-art models with and without the adoption of the proposed SAM. 

Section~\ref{subsec: Insights_SOA} shows the performance of state-of-the-art methods evaluated on the introduced Semantic Recall metric. In Section~\ref{sec:reduced_data} we present the significantly better performance achieved at retrieval when using considerably less training data compared to current state-of-the-art models. Section~\ref{sec:soa} showcases several state-of-the-art models with and without the adoption of our adaptive margin formulation. Finally, Section~\ref{section:ablation} presents the effects of employing the original triplet formulation, different values of a temperature parameter $\tau$ and different sampling strategies. 


In all our experiments, we employ publicly available code from the authors and train the models from scratch according to the original strategy and hyper-parameters. 
In order to perform a fair comparison, we do not use ensembles in our experiments. We employed CIDEr to assess the similarity between samples at training time ($\phi$). With the purpose of avoid training and evaluating on similar metrics, we employ SPICE when NSC is used as an evaluation metric. For the implementation details of each of our adaptive margin formulation we kindly ask the reader to refer to the supplementary material section.

\subsection{Insights on State-of-the-Art Retrieval}
\label{subsec: Insights_SOA}
\begin{figure}
    \centering
    \includegraphics[width=\linewidth]{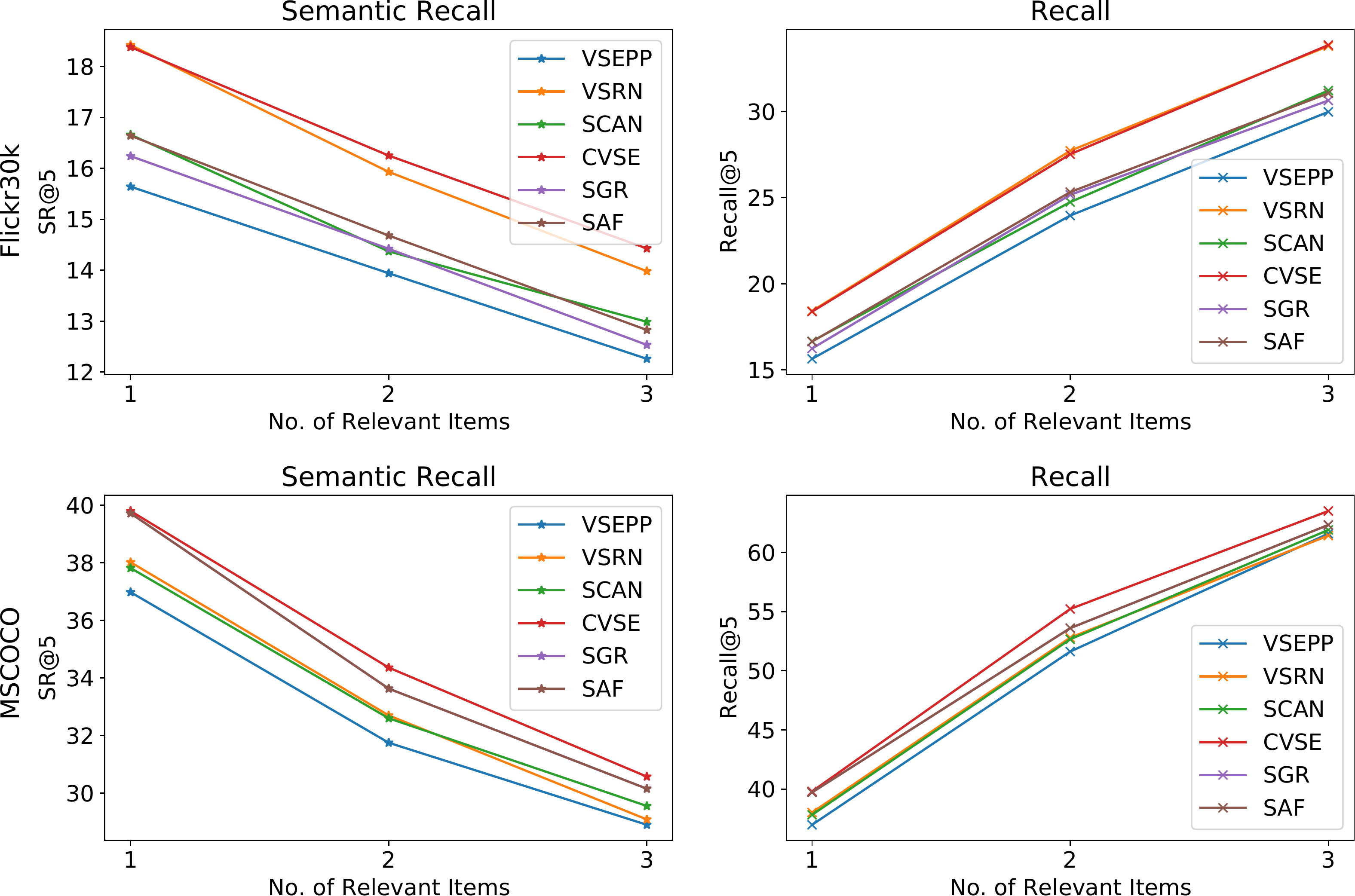}
    \caption{Text-to-Image Top-5 retrieved results evaluated with Recall and the the presented Semantic Recall for Non-GT items.}
    \label{fig:reality_check}
\end{figure}

In this section, we compare the behaviour of existing systems by evaluating them on the newly proposed metrics. We evaluate the following state-of-the-art models: VSE++~\cite{faghri2017vse++}, SCAN~\cite{lee2018stacked}, VSRN~\cite{li2019visual}, CVSE~\cite{wang2020consensus}, SGR and SAF~\cite{diao2021similarity}. The experiment depicting the top-$5$ text-to-image retrieval scores for non-ground truth relevant items is shown in Figure~\ref{fig:reality_check}. The scores shown are in terms of Recall and Semantic Recall at a cut-off point $5$. Figures showcasing the remaining scenarios can be found in the supplementary material. 

It is worth noting in Figure~\ref{fig:reality_check} that according to the recall (R@5), the models have a steady raise in recall scores as the number of relevant images $m$ increase. However, the opposite effect is found when the models are evaluated with the previously introduced Semantic Recall (SR) formulation. The behaviour of the models according to these two metrics seems to have an inversely proportional relation. The reason is due to the different definition between $R$ and $R^V$. Merely evaluating the models on the first correctly retrieved item does not provide a complete landscape of its performance. Instead, our formulation shows that the models tend to have a decreasing score when more relevant items are considered.
Furthermore, we observe that the big difference in numbers between models seems to diminish when we increase the relevant items for both metrics. Our conclusion is that the performance boost we obtain in the literature is not reflected well on the non-GT relevant items, suggesting that the generalization power of the models is overestimated.

\subsection{Reduced Data Scenario}
\label{sec:reduced_data}
 
 \definecolor{gray}{RGB}{210,210,210}
\newcolumntype{a}{>{\columncolor{gray}}c}

\begin{table*}[!h]
\begin{center}
\scriptsize

\begin{tabular}{|l|ccc|ccc|c|ccc|ccc|c}

\multicolumn{15}{c}{\textbf{Flickr30K}} \\ \hline
\textbf{} & \multicolumn{7}{c|}{\textbf{Recall}} & \multicolumn{7}{c|}{\textbf{Normalized Cumulative Semantic Score}} \\ \hline
\textbf{Method} & \multicolumn{3}{c|}{\textbf{I2T}} & \multicolumn{3}{c|}{\textbf{T2I}} & \multicolumn{1}{l|}{} & \multicolumn{3}{c|}{\textbf{I2T}} & \multicolumn{3}{c|}{\textbf{T2I}} & \multicolumn{1}{l|}{} \\ \cline{2-15} 
 & \textbf{R@1} & \textbf{R@5} & \textbf{R@10} & \textbf{R@1} & \textbf{R@5} & \textbf{R@10} & \textbf{Rsum} & \textbf{N@1} & \textbf{N@5} & \textbf{N@10} & \textbf{N@1} & \textbf{N@5} & \textbf{N@10} & \multicolumn{1}{l|}{\textbf{Nsum}} \\ \hline

VSRN\cite{li2019visual}$\dagger$                            & 68.1          & 88.4          & 93.9          & 51.6          & 78.3          & \multicolumn{1}{c|}{85.8}          & 466.1                          & \textbf{60.3} & \textbf{62.9} & 62.9           & 63.2          & 58.8          & \multicolumn{1}{c|}{55.0}          & \multicolumn{1}{c|}{363.4}                             \\
VSRN+\scriptsize{SAM}                         & \textbf{68.4} & \textbf{89.7} & \textbf{94.8} & \textbf{52.4} & \textbf{78.7} & \multicolumn{1}{c|}{\textbf{86.6}} & \textbf{470.6}                 & 60.2          & 62.7          & \textbf{63.1}  & \textbf{64.1} & \textbf{59.4} & \multicolumn{1}{c|}{\textbf{55.7}} & \multicolumn{1}{c|}{\textbf{365.2}}                   \\ \hline
CVSE \cite{wang2020consensus}$\dagger$                            & 68.6          & 87.7          & 92.7          & 53.2          & 81.1          & \multicolumn{1}{c|}{88.3}          & 471.6                          & 59.0          & 63.5          & 63.1          & 64.1          & 59.6           & \multicolumn{1}{c|}{\textbf{55.5}}           & \multicolumn{1}{c|}{364.9}                             \\
CVSE +  \tiny{SAM}                         & \textbf{70.0}   & \textbf{89.2} & \textbf{93.1} & \textbf{55.0} & \textbf{82.6} & \multicolumn{1}{c|}{\textbf{89.0}} & \textbf{478.9}                 & \textbf{59.6} & \textbf{64.6} & \textbf{64.2} & \textbf{65.5} & \textbf{59.8} & \multicolumn{1}{c|}{55.3} & \multicolumn{1}{c|}{\textbf{387.9}}                    \\ 
\hline
SGR\cite{diao2021similarity}$\dagger$                            & 74.4          & \textbf{92.9} & 96.3          & 55.8          & 81.1          & \multicolumn{1}{c|}{87.9}          & 488.4                          & 68.1           & 65.6          & 63.8          & 66.0          & 58.6          & \multicolumn{1}{c|}{54.5}          & \multicolumn{1}{c|}{376.7}                             \\
SGR+\scriptsize{SAM}                          & \textbf{75.9} & 92.4          & \textbf{96.6} & \textbf{57.6} & \textbf{83.1} & \multicolumn{1}{c|}{\textbf{89.7}} & \textbf{495.3}                 & \textbf{69.4} & \textbf{66.2} & \textbf{64.0} & \textbf{67.5} & \textbf{59.2} & \multicolumn{1}{c|}{\textbf{55.0}} & \multicolumn{1}{c|}{\textbf{381.4}}

\\ \hline
\multicolumn{15}{c}{\textbf{MSCOCO 1K}} \\ \hline
\textbf{} & \multicolumn{7}{c|}{\textbf{Recall}} & \multicolumn{7}{c|}{\textbf{Normalized Cumulative Semantic Score}} \\ \hline
\textbf{Method} & \multicolumn{3}{c|}{\textbf{I2T}} & \multicolumn{3}{c|}{\textbf{T2I}} & \multicolumn{1}{l|}{} & \multicolumn{3}{c|}{\textbf{I2T}} & \multicolumn{3}{c|}{\textbf{T2I}} & \multicolumn{1}{l|}{} \\ \cline{2-15} 
 & \textbf{R@1} & \textbf{R@5} & \textbf{R@10} & \textbf{R@1} & \textbf{R@5} & \textbf{R@10} & \textbf{Rsum} & \textbf{N@1} & \textbf{N@5} & \textbf{N@10} & \textbf{N@1} & \textbf{N@5} & \textbf{N@10} & \multicolumn{1}{l|}{\textbf{Nsum}} \\ \hline
VSRN\cite{li2019visual}$\dagger$ & 72.4 & \textbf{94.7} & \textbf{97.8} & 61.2 & 89.3 & \textbf{94.9} & 510.3 & 68.3 & 72.1 & \textbf{68.2} & 74.4 & \textbf{71.1} & 66.4 & \multicolumn{1}{c|}{420.6} \\
VSRN+\scriptsize{SAM} & \textbf{74.6} & 93.6 & 97.5 & \textbf{61.5} & \textbf{89.6} & \textbf{94.9} & \textbf{511.7} & \textbf{69.3} & \textbf{72.2} & 68.1 & \textbf{74.5} & 70.9 & \textbf{66.5} & \multicolumn{1}{c|}{\textbf{421.5}} \\ \hline
CVSE\cite{wang2020consensus}$\dagger$ & 77.0 & 94.2 & 97.3 & 64.3 & 91.1 & 95.9 & 519.8 & 69.7 & 73.3 & 69.3 & 76.2 & 71.4 & 67.1 & \multicolumn{1}{c|}{427.2} \\
CVSE+\scriptsize{SAM} & \textbf{79.8} & \textbf{95.1} & \textbf{97.7} & \textbf{67.0} & \textbf{93.0} & \textbf{97.3} & \textbf{529.9} & \textbf{71.8} & \textbf{76.3} & \textbf{71.0} & \textbf{78.6} & \textbf{72.9} & \textbf{69.1} & \multicolumn{1}{c|}{\textbf{439.6}} \\ 
\hline
SGR\cite{diao2021similarity}$\dagger$ & 79.9 & \textbf{97.4} & 98.3 & 63.2 & \textbf{90.5} & 95.4 & 524.7 & \textbf{74.7} & \textbf{73.1} & \textbf{67.9} & 76.1 & 70.7 & 67.2 & \multicolumn{1}{c|}{\textbf{429.9}} \\
SGR+\scriptsize{SAM} & \textbf{80.7} & 97.2 & \textbf{98.6} & \textbf{63.8} & \textbf{90.5} & \textbf{95.9} & \textbf{526.7} & 73.2 & 72.9 & 67.8 & \textbf{76.2} & \textbf{70.9} & \textbf{67.4} & \multicolumn{1}{c|}{428.5} \\ \hline

\multicolumn{15}{c}{\textbf{MSCOCO 5K}} \\ \hline
\textbf{} & \multicolumn{7}{c|}{\textbf{Recall}} & \multicolumn{7}{c|}{\textbf{Normalized Cumulative Semantic Score}} \\ \hline
\textbf{Method} & \multicolumn{3}{c|}{\textbf{I2T}} & \multicolumn{3}{c|}{\textbf{T2I}} & \multicolumn{1}{l|}{} & \multicolumn{3}{c|}{\textbf{I2T}} & \multicolumn{3}{c|}{\textbf{T2I}} & \multicolumn{1}{l|}{} \\ \cline{2-15} 
 & \textbf{R@1} & \textbf{R@5} & \textbf{R@10} & \textbf{R@1} & \textbf{R@5} & \textbf{R@10} & \textbf{Rsum} & \textbf{N@1} & \textbf{N@5} & \textbf{N@10} & \textbf{N@1} & \textbf{N@5} & \textbf{N@10} & \multicolumn{1}{l|}{\textbf{Nsum}} \\ \hline

VSRN\cite{li2019visual}$\dagger$ & 48.4 & 78.9 & \textbf{87.9} & 37.2 & 67.9 & \textbf{79.6} & 399.9 & 55.8 & 58.6 & 61.2 & 60.2 & 63.4 & 62.8 & \multicolumn{1}{c|}{362.2} \\
VSRN+\scriptsize{SAM} & \textbf{49.1} & \textbf{79.0} & 87.4 & \textbf{37.5} & \textbf{68.1} & 79.5 & \textbf{400.6} & \textbf{56.4} & \textbf{58.8} & \textbf{61.7} & \textbf{60.6} & \textbf{63.5} & \textbf{62.9} & \multicolumn{1}{c|}{\textbf{363.9}} \\ \hline
CVSE\cite{wang2020consensus}$\dagger$ & 53.1 & 79.6 & 88.0 & 40.5 & 72.2 & 83.1 & 416.5 & 57.1 & 61.1 & 63.2 & 62.2 & 64.4 & 63.4 & \multicolumn{1}{c|}{371.4} \\
CVSE+\scriptsize{SAM} & \textbf{56.4} & \textbf{82.4} & \textbf{90.1} & \textbf{42.3} & \textbf{73.9} & \textbf{84.5} & \textbf{429.6} & \textbf{59.2} & \textbf{63.0} & \textbf{64.5} & \textbf{63.8} & \textbf{65.3} & \textbf{64.4} & \multicolumn{1}{c|}{\textbf{380.2}} \\ 

\hline
SGR\cite{diao2021similarity}$\dagger$ & \textbf{56.0} & \textbf{83.3} & 90.7 & 40.1 & 69.3 & 80.2 & 419.6 & \textbf{60.4} & 59.1 & \textbf{60.4} & 61.7 & \textbf{62.5} & \textbf{61.6} & \multicolumn{1}{c|}{\textbf{366.0}} \\
SGR+\scriptsize{SAM} & 55.7 & 83.2 & \textbf{91.2} & \textbf{40.5} & \textbf{69.7} & \textbf{80.5} & \textbf{420.8} & 59.5 & \textbf{59.3} & \textbf{60.4} & \textbf{62.0} & 62.4 & 61.4 & \multicolumn{1}{c|}{365.0}\\ \hline

\end{tabular}

\end{center}
\caption{Comparison of retrieval results of the original VSRN, CVSE and SGR models with and without the proposed SAM. Results are  depicted in terms of Recall@K (R@K) and Normalized Cumulative Semantic Score (N@K). The column Rsum and Nsum is the summation of the overall retrieval scores in image-to-text and text-to-image for Recall and NCS respectively. The $\dagger$ depicts that models are trained with the publicly available code released by the original authors.}

\label{tab:full_sota_results}
\end{table*}
We hypothesize that our adaptive margin formulation based on CIDEr is better equipped to deal with scarce training data scenarios, as it can better exploit the semantics over the whole data. More explicitly, we set aside a similar proportion of training samples from Flickr30k ($29,000$) and MSCOCO ($113,287$). In Flickr30K we employed 10\% and 25\% of the training set, resulting in $2,900$ and $7,250$ samples respectively. In the case of MSCOCO we employed 3\% and 5\% of the training set, thus yielding $3,398$ and $5,664$ data points. We evaluate all the models on the standard $1$K test set split of each dataset. We employ two state-of-the-art methods, CVSE~\cite{wang2020consensus} and SGR\cite{diao2021similarity} for experimentation. Similar to the previous section, all the experiments are performed with public available code as described by the authors disregarding the adoption or not of our formulation. The results these experiments can be found in Table~\ref{tab:minidata_results}. 

In the 10\% data scenario of Flickr30k, CVSE with SAM achieves almost $3$ times the performance when compared to the original model. It is paramount to note that by the adoption of our formulation SGR achieves an enormous improvement. On the other hand, the original SGR model is barely capable to learn useful information due to the bigger number of parameters compared to CVSE. 

As more data is used on each scenario, the original models tend to improve in performance and the retrieval gap decreases. Results in Flickr30k tend to be stronger when the adopting the proposed SAM. This is due to the significantly higher descriptive nature of captions found in Flickr30K training dataset compared to the less granular ones found in MSCOCO.

The significant improvement in scarce training data, also translates into an increased rate of convergence. By employing an adaptive margin with CIDEr, a model exploits a strong guiding cue about the semantic space to be learned according to weighted n-gram statistics.

\subsection{Comparison with State-of-the-Art}
\label{sec:soa}

The results obtained by comparing state-of-the-art methods with and without our formulated 
SAM are shown in Table~\ref{tab:full_sota_results}. 
First, by incorporating SAM, calculated from an image captioning metric into a state-of-the-art pipeline, a boost in recall is obtained. A similar effect is achieved in most of the models when they are also evaluated with the proposed NCS metric. 

Second, both depicted metrics have a strong degree of correlation, however, obtaining an improvement in recall does not necessarily translate in an increase in NCS. This effect can be observed in particular with the MSCOCO dataset with the SGR model.

This is due to the fact that Recall and NCS are inherently different metrics that provide complementary information. Recall shows how well a model ranks a single image or sentence labeled as relevant. Whereas, the NCS shows what is the degree of semantics captured by a model at a cut-off point k. Therefore, an increase in Recall or NCS should not necessarily be treated as equally significant.

Third, it is evident that a greater improvement is achieved on Flickr30k than in MSCOCO. All of our models on Flickr30k perform better than the baselines on every metric, while in MSCOCO, the boost attained is more conservative. The reason is that captions of Flickr30k are more detailed and longer compared to the ones in MSCOCO, which are shorter and less specific. This difference in the nature of the captions allows CIDEr to provide a more precise and discriminative margin per sample in intricate captions, due to the CIDEr formulation which relies on a weighted tf-idf n-gram matching.  


Finally, it is important to note that while the Recall score increases as the cut-off point increases, in our proposed NCS metric this behaviour is not present. The NCS shows the normalized capability of a model to capture the greatest amount of semantic similarity on a specific cut-off point.





\subsection{Effect of Temperature and Sampling}
\label{section:ablation}

In this section, we study the effect of: the temperature parameter $\tau$, sampling techniques, and whether the original triplet is kept or only a SAM is employed. Several sampling techniques are explored to find the negative items in our SAM formulation, namely random (RS), hard negative (HN) and soft negative(SN). In HN, the negative item in each triplet is selected as the closest to the anchor in a batch~\cite{faghri2017vse++}. We refer to random sampling when a negative item is randomly picked in a batch. SN refers to picking the furthest negative item to the anchor within the batch. We investigate the effect of these parameters employing CVSE~\cite{wang2020consensus} model as a baseline.
The majority of the best performing models obtained were employing a Soft Negative (SN) sampling, thus we provide the results on both datasets in Table~\ref{tab:ablation_sn}. The results of the effect of Random Sampling (RS) and Hard Negative (HN) sampling in Flickr30k are shown in Table~\ref{tab:ablation_rs_hn}. In both tables, we provide the sum of the Recall and NCS metrics at the top $1$, $5$ and $10$ in image-to-text and text-to-image scenarios. When the NCS is employed, we show two variations. One, by preserving the GT images labeled as relevant and the second one by removing only the GT images, denoted with the acronym N in Table~\ref{tab:ablation_sn} and~\ref{tab:ablation_rs_hn}. We ask the reader to refer to the supplementary material section for more details.

\definecolor{gray}{RGB}{210,210,210}
\newcolumntype{a}{>{\columncolor{gray}}c}

\begin{table}[!h]
\begin{center}
\footnotesize
\setlength{\tabcolsep}{3pt}
\begin{tabular}{c|c|c|ccc|ccc}
{\multirow{2}{*}{\textbf{$\tau$}}} & {\multirow{2}{*}{\textbf{S}}} & {\multirow{2}{*}{\textbf{T}}} & \multicolumn{3}{c|}{\textbf{F30K}} & \multicolumn{3}{c}{\textbf{MSCOCO-1K}} \\ \cline{4-9} 
 &  &  & \multicolumn{1}{l}{\textbf{Nsum}} & \multicolumn{1}{l}{\textbf{Nsum(N)}} & \multicolumn{1}{l|}{\textbf{Rsum}} & \multicolumn{1}{l}{\textbf{Nsum}} & \multicolumn{1}{l}{\textbf{Nsum(N)}} & \multicolumn{1}{l}{\textbf{Rsum}} \\ \hline
3 & SN & \cmark & \textbf{371.29} & 257.42 & \textbf{479.1} & 429.41 & 312.2 & 517.9 \\
3 & SN & \xmark & 369.73 & 258.12 & 476.5 & 427.32 & \textbf{313.06} & 515 \\
5 & SN & \cmark & 369.97 & 257.21 & 477.5 & 429.74 & 311.53 & 520.5 \\
5 & SN & \xmark & \textbf{371.31} & \textbf{258.75} & 478.2 & 428.2 & 312.89 & 518 \\
10 & SN & \cmark & 369.74 & 257.64 & 477.2 & 429.09 & 306.51 & \textbf{521.3} \\
10 & SN & \xmark & 370.17 & 257.85 & 475.8 & \textbf{429.88} & 309.27 & 520.6\\
\hline
\end{tabular}

\end{center}
\caption{Experiments of the effect of ($\tau$), soft negative (SN) sampling and the whether the original triplet is kept (\cmark) or only our formulation is employed (\xmark). The acronym Nsum(N) refers that GT elements have been removed.}

\label{tab:ablation_sn}
\end{table}

\definecolor{gray}{RGB}{210,210,210}
\newcolumntype{a}{>{\columncolor{gray}}c}

\begin{table}[!h]
\begin{center}
\footnotesize

\begin{tabular}{c|c|c|ccc}
\multirow{2}{*}{\textbf{$\tau$}} & \multirow{2}{*}{\textbf{S}} & \multirow{2}{*}{\textbf{T}} & \multicolumn{3}{c}{\textbf{F30K}} \\ \cline{4-6} 
 &  &  & \multicolumn{1}{l}{\textbf{Nsum}} & \multicolumn{1}{l}{\textbf{Nsum(N)}} & \multicolumn{1}{l}{\textbf{Rsum}} \\ \hline
3 & RS & \cmark & 355.87 & 257.93 & 460.1 \\
3 & RS & \xmark & 344.44 & 257.67 & 441.7 \\
5 & RS & \cmark & 367.02 & 258.25 & 473.4 \\
5 & RS & \xmark & 363.27 & \textbf{259.83} & 468.3 \\
10 & RS & \cmark & \textbf{370.09} & 257.42 & \textbf{478.7} \\
10 & RS & \xmark & 365.72 & 258.82 & 471.2 \\ \hline
3 & HN & \cmark & 338.94 & 249.04 & 435.1 \\
3 & HN & \xmark & 344.19 & \textbf{258.27} & 439.5 \\
5 & HN & \cmark & \textbf{369.94} & 257.97 & \textbf{478.8} \\
5 & HN & \xmark & 351.68 & 257.64 & 450.4 \\
10 & HN & \cmark & 369.04 & 256.85 & 477.2 \\
10 & HN & \xmark & 351.18 & 257.83 & 448.6 \\ \hline
\end{tabular}
\end{center}
\caption{Experiments of the effect of ($\tau$), random (RS) and hard negative (HN) sampling. The third column (T) shows whether the original triplet is kept (\cmark) or only our formulation is employed (\xmark). The acronym Nsum(N) refers that GT elements have been removed.}

\label{tab:ablation_rs_hn}
\end{table}

Initially it is important to notice that improvements on the recall score do not necessarily go in hand with better scores at NCS. This can be seen in the MSCOCO-1K results between fifth row and first row in Table~\ref{tab:ablation_sn}. In these experiments, we obtain a score of $521$ on Rsum and $429$ on NCS sum in the fifth row. Comparing it to the first row, there is a 4\% drop on recall, however the score of $429$ remains on NCS. Although NCS and Recall are correlated, they provide different information of our models. 

In general, we obtain our best NCS scores when the temperature parameter $\tau$ is increased to $10$. 
The smaller margin gives the model more freedom in shaping the space on where to project the data points. By increasing the margin, we restrict the models on where to project the positive and negative samples, resulting in a drop in NCS and Recall. However, a trade-off between NCS scores on GT and non-GT items exists. The increase in the margin (lower values in $\tau$) seems to improve the results on non-GT items, this is especially evident on MSCOCO.  We discover that on average, we obtain the best results with SN. Regarding the usage of the original triplet formulation, we notice that it is complementary to SAM since each one focuses on learning a different task. The hard negative focuses solely on GT samples, while SAM learns to measure the degree of similarity.




\section{Conclusion}
In this work, we highlight the challenges stemming from the lack of annotations in the task of image-text matching. Inspired by image captioning metrics, we present a formulation that addresses the many-to-many mapping problem between images and captions. The introduced metric, namely Normalized Cumulative Semantic Score (NCS), shows higher degree of semantic correlation to human judgement compared to the standard Recall. 
Additionally, we show a comprehensive set of experiments that considers the usage of IC metrics to learn an adaptive margin. The incorporation of such margin yields a big improvement in scenarios when training data is scarce (eg semi-supervised learning), as well as increasing the semantics of the retrieved non-GT items. 


{\small
\bibliographystyle{ieee_fullname}
\bibliography{egbib}
}

\end{document}


\title{Supplementary Material\\Paper ID 499: Is An Image Worth Five Sentences? \\ 
A New Look into Semantics for Image-Text Matching}
\maketitle

\section{Implementation Details}
In this section, we describe the hyper-parameters and the training procedure used to obtain the models shown in the main paper for the reduced data scenario (Table 2 - main paper) and the state-of-the-art comparison (Table 3 - main paper).
Specifically, for each model we employ the training procedure described in the original paper. Each of the models use the $36$ most confident regions obtained by a Faster R-CNN~\cite{girshick2015fast} from the object detector proposed by~\cite{anderson2017bottom}. The visual features used by each model are the same in all cases. The best model is selected according to the sum of NCS with CIDEr as a similarity metric on the validation set. It is important to note that NCS with SPICE is the one we employ for evaluation.

For the model VSRN~\cite{li2019visual} + SAM, we start with the public pretrained weights. We add our SAM loss function and re-train the model for $30$ epochs. For Flickr30K we employ a soft negative sampling, with a temperature parameter $\tau=3$ and a weight on the SAM triplet loss of $20$. When training on MSCOCO, we employ a soft negative sampling, with a temperature parameter $\tau=10$ and a weight on the SAM triplet loss of $5$. In both datasets, the original triplet loss is kept alongside with our SAM formulation. As in the original paper, the word embedding size is $300$-d and the dimension of the final joint embedding space is $2048$-d. The mini-batch size employed is $128$. At training, the Adam optimizer~\cite{kingma2014adam} is used. The original triplet loss margin is $0.2$. The model is trained for $30$ epochs, with a learning rate of $0.0002$ for the initial $15$ epochs and is divided by 10 for the remaining $15$ epochs. 

In the CVSE~\cite{wang2020consensus} + SAM model, we train the model from scratch alongside with the proposed SAM loss function as follows. In Flickr30K, we employ a random sampling strategy with a temperature parameter $\tau=7$ and a weight on the SAM triplet loss of $5$ alongside with the original triplet. In MSCOCO, we employ a soft negative sampling strategy, with a temperature parameter $\tau=5$ and a weight on the SAM triplet loss set to $5$, and only our SAM triplet loss is used. Following the original CVSE model, the joint space dimension is $1024$-d. The consensus exploitation is performed with a $300$-d GloVe~\cite{pennington2014glove} representation. The loss formulation contains the following weights for each term are kept, $\lambda_1=3, \lambda_2=5, \lambda_3=1, \lambda_4=2$. The mini-batch size employed is $128$. At training, the Adam optimizer~\cite{kingma2014adam} is used.  The original triplet loss margin is $0.2$. The model is trained for $30$ epochs, with a learning rate of $0.0002$ for the initial $15$ epochs and is divided by 10 for the remaining $15$ epochs. 

Finally, for the SGR~\cite{diao2021similarity} + SAM model, we train the model from scratch in both datasets. In Flickr30K, we employ a random sampling strategy, with a temperature parameter $\tau=10$ and a weight on the SAM triplet loss of $10$. In MSCOCO, we use a random sampling strategy, with a temperature parameter $\tau=5$ and a weight on the SAM triplet loss of $5$. In both datasets, the original triplet is kept alongside with our SAM triplet. Training of the original model is performed as described by the authors. The word embedding size is $300$-d and the number of hidden states is $1024$-d. The dimension of the similarity representation is $256$. The original triplet loss margin is $0.2$. The number of reasoning steps is $3$. The initial learning rate is set to $0.0002$ for $10$ epochs and is decreased by a $10$ on the final $10$ epochs on MSCOCO. For Flickr30K, the initial learning rate is kept for $30$ epochs and it decays by 0.1 for the next $10$ epochs.

\section{Effect of Temperature and Sampling}

In this section, we extend the results of the effect of setting different temperature values $\tau$ and sampling strategies. In all the experiments in this section, CVSE~\cite{wang2020consensus} is used.
We evaluate the impact of these parameters on Flickr30K on Table~\ref{tab:f30k_ablation1} and removing GT items (non-GT) on Table~\ref{tab:f30k_ablation2}. When calculating the results on MSCOCO 1K, we keep the soft negative (SN) sampling strategy while the impact of different values of $\tau$ and the usage of the original triplet is measured. Table~\ref{tab:coco_ablation1} shows the results with the inclusion of GT and Table~\ref{tab:coco_ablation2} depicts the results obtained with non-GT relevant items.

\definecolor{gray}{RGB}{210,210,210}
\newcolumntype{a}{>{\columncolor{gray}}c}

\begin{table*}[!h]
\begin{center}
\scriptsize

\begin{tabular}{|c|c|c|ccc|ccc|c|ccc|ccc|c|}
\hline
\multirow{3}{*}{\textbf{$\tau$}} & \multirow{3}{*}{\textbf{S}} & \multirow{3}{*}{\textbf{T}} & \multicolumn{7}{c|}{\textbf{Recall}} & \multicolumn{7}{c|}{\textbf{Normalized Cumulative Semantic Recall}} \\ \cline{4-17} 
 &  &  & \multicolumn{3}{c|}{\textbf{I2T}} & \multicolumn{3}{c|}{\textbf{T2I}} & \multicolumn{1}{l|}{} & \multicolumn{3}{c|}{\textbf{I2T}} & \multicolumn{3}{c|}{\textbf{T2I}} & \multicolumn{1}{l|}{} \\ \cline{4-17} 
 &  &  & \textbf{R@1} & \textbf{R@5} & \textbf{R@10} & \textbf{R@1} & \textbf{R@5} & \textbf{R@10} & \textbf{Rsum} & \textbf{N@1} & \textbf{N@5} & \textbf{N@10} & \textbf{N@1} & \textbf{N@5} & \textbf{N@10} & \multicolumn{1}{l|}{\textbf{Nsum}} \\ \hline
3 & SN &\cmark& 69.8 & 87.8 & 93.1 & 55.5 & 83.2 & 89.7 & 479.1 & 59.6 & 65.4 & 64.6 & 65.9 & 60.1 & 55.8 & 371.3 \\
3 & SN &\xmark& 70.0 & 88.1 & 92.6 & 54.1 & 82.4 & 89.3 & 476.5 & 60.0 & 64.7 & 64.5 & 64.8 & 60.0 & 55.8 & 369.7 \\
5 & SN &\cmark& 69.7 & 88.4 & 93.1 & 54.5 & 82.4 & 89.4 & 477.5 & 59.9 & 65.2 & 64.4 & 65.2 & 59.7 & 55.5 & 370.0 \\
5 & SN &\xmark& 70.4 & 88.2 & 92.5 & 55.1 & 82.6 & 89.4 & 478.2 & 60.3 & 65.6 & 64.3 & 65.5 & 59.8 & 55.9 & 371.3 \\
10 & SN &\cmark& 69.6 & 89.1 & 92.3 & 54.9 & 82.2 & 89.1 & 477.2 & 60.0 & 65.4 & 64.2 & 65.3 & 59.6 & 55.3 & 369.7 \\
10 & SN &\xmark& 70.6 & 86.5 & 92.9 & 54.1 & 82.3 & 89.4 & 475.8 & 60.9 & 65.1 & 64.3 & 64.7 & 59.6 & 55.7 & 370.2 \\ \hline
3 & RS &\cmark& 65.4 & 85.7 & 91.6 & 50.9 & 79.1 & 87.4 & 460.1 & 56.3 & 61.1 & 61.5 & 62.1 & 59.1 & 55.9 & 355.9 \\
3 & RS &\xmark& 60.9 & 84.3 & 89.4 & 46.9 & 76.0 & 84.2 & 441.7 & 53.8 & 58.9 & 59.7 & 59.0 & 58.0 & 55.1 & 344.4 \\
5 & RS &\cmark& 68.3 & 88.1 & 92.7 & 54.0 & 81.5 & 88.8 & 473.4 & 59.1 & 63.9 & 63.8 & 64.7 & 59.6 & 56.0 & 367.0 \\
5 & RS &\xmark& 67.5 & 87.4 & 92.1 & 52.9 & 80.3 & 88.1 & 468.3 & 58.1 & 63.0 & 63.0 & 63.8 & 59.4 & 56.0 & 363.3 \\
10 & RS &\cmark& 70.7 & 87.8 & 93.7 & 54.7 & 82.4 & 89.4 & 478.7 & 60.9 & 64.8 & 64.0 & 65.2 & 59.8 & 55.4 & 370.1 \\
10 & RS &\xmark& 68.0 & 87.9 & 93.0 & 52.8 & 81.2 & 88.3 & 471.2 & 59.5 & 63.9 & 63.5 & 63.8 & 59.5 & 55.5 & 365.7 \\\hline
3 & HN &\cmark& 59.6 & 81.4 & 89.3 & 44.9 & 75.6 & 84.3 & 435.1 & 54.1 & 57.4 & 58.9 & 56.9 & 57.2 & 54.4 & 338.9 \\
3 & HN &\xmark& 60.5 & 85.8 & 91.5 & 44.6 & 73.9 & 83.2 & 439.5 & 55.5 & 58.7 & 59.6 & 57.8 & 57.5 & 55.0 & 344.2 \\
5 & HN &\cmark& 71.1 & 88.7 & 92.7 & 54.4 & 82.6 & 89.3 & 478.8 & 60.6 & 64.6 & 64.1 & 65.1 & 59.8 & 55.7 & 369.9 \\
5 & HN &\xmark& 62.9 & 86.3 & 92.7 & 47.4 & 76.7 & 84.4 & 450.4 & 57.6 & 60.1 & 60.9 & 59.6 & 58.2 & 55.3 & 351.7 \\
10 & HN &\cmark& 70.0 & 88.3 & 92.6 & 54.7 & 82.4 & 89.2 & 477.2 & 60.2 & 64.5 & 64.0 & 65.1 & 59.8 & 55.5 & 369.0 \\
10 & HN &\xmark& 63.5 & 85.1 & 92.0 & 47.2 & 76.3 & 84.5 & 448.6 & 57.2 & 60.6 & 60.8 & 59.5 & 58.1 & 55.0 & 351.2 \\ \hline
\end{tabular}

\end{center}
\caption{Experiments on Flickr30K regarding the effect of ($\tau$), soft (SN), random (RS) and hard negative (HN) sampling . The third column (T) shows whether the original triplet is kept (\cmark) or only our formulation is employed (\xmark). For all the experiments shown, CVSE\cite{wang2020consensus} is employed. Results are depicted in terms of Recall@K (R@K) and Normalized Cumulative Semantic Score (N@K).}

\label{tab:f30k_ablation1}
\end{table*}

\definecolor{gray}{RGB}{210,210,210}
\newcolumntype{a}{>{\columncolor{gray}}c}

\begin{table*}[!h]
\begin{center}
\scriptsize

\begin{tabular}{|c|c|c|ccc|ccc|c|}
\hline
\multirow{3}{*}{\textbf{$\tau$}} & \multirow{3}{*}{\textbf{S}} & \multirow{3}{*}{\textbf{T}} & \multicolumn{7}{c|}{\textbf{Normalized Cumulative Semantic Recall (Non-GT)}} \\ \cline{4-10} 
 &  &  & \multicolumn{3}{c|}{\textbf{I2T}} & \multicolumn{3}{c|}{\textbf{T2I}} & \multicolumn{1}{l|}{} \\ \cline{4-10} 
 &  &  & \textbf{N@1} & \textbf{N@5} & \textbf{N@10} & \textbf{N@1} & \textbf{N@5} & \textbf{N@10} & \multicolumn{1}{l|}{\textbf{Nsum}} \\ \hline
3 & SN &\cmark& 40.9 & 42.4 & 43.1 & 42.7 & 44.1 & 44.4 & 257.5 \\
3 & SN &\xmark& 41.2 & 42.3 & 43.1 & 42.8 & 44.2 & 44.5 & 258.1 \\
5 & SN &\cmark& 41.4 & 42.3 & 42.8 & 42.8 & 43.9 & 44.1 & 257.2 \\
5 & SN &\xmark& 41.5 & 42.6 & 43.4 & 42.7 & 44.0 & 44.6 & 258.8 \\
10 & SN &\cmark& 42.3 & 42.3 & 42.8 & 42.6 & 43.7 & 43.9 & 257.6 \\
10 & SN &\xmark& 41.3 & 42.5 & 43.0 & 42.7 & 44.0 & 44.4 & 257.8 \\ \hline
3 & RS &\cmark& 40.4 & 42.1 & 43.1 & 42.6 & 44.6 & 45.1 & 257.9 \\
3 & RS &\xmark& 40.7 & 42.2 & 43.0 & 42.4 & 44.4 & 45.0 & 257.7 \\
5 & RS &\cmark& 40.9 & 42.6 & 43.1 & 42.5 & 44.2 & 45.0 & 258.3 \\
5 & RS &\xmark& 41.0 & 43.1 & 43.5 & 42.6 & 44.5 & 45.1 & 259.8 \\
10 & RS &\cmark& 41.7 & 42.6 & 42.8 & 42.6 & 43.8 & 43.9 & 257.4 \\
10 & RS &\xmark& 41.5 & 43.0 & 43.3 & 42.5 & 44.1 & 44.4 & 258.8 \\
\hline
3 & HN &\cmark& 38.4 & 40.5 & 41.4 & 41.1 & 43.5 & 44.3 & 249.0 \\
3 & HN &\xmark& 41.2 & 41.9 & 42.7 & 42.8 & 44.4 & 45.3 & 258.3 \\
5 & HN &\cmark& 41.2 & 42.4 & 43.0 & 42.9 & 44.0 & 44.4 & 258.0 \\
5 & HN &\xmark& 40.1 & 42.3 & 43.0 & 42.6 & 44.4 & 45.2 & 257.7 \\
10 & HN &\cmark& 41.0 & 42.4 & 43.0 & 42.3 & 43.9 & 44.1 & 256.8 \\
10 & HN &\xmark& 41.2 & 42.6 & 43.0 & 42.1 & 44.3 & 44.6 & 257.8 \\ \hline
\end{tabular}

\end{center}
\caption{Experiments on Flickr30K (Non-GT) regarding the effect of ($\tau$), soft (SN), random (RS) and hard negative (HN) sampling . The third column (T) shows whether the original triplet is kept (\cmark) or only our formulation is employed (\xmark). For all the experiments shown, CVSE\cite{wang2020consensus} is employed. Results are depicted in terms of Recall@K (R@K) and Normalized Cumulative Semantic Score (N@K).}

\label{tab:f30k_ablation2}
\end{table*}

\definecolor{gray}{RGB}{210,210,210}
\newcolumntype{a}{>{\columncolor{gray}}c}

\begin{table*}[!h]
\begin{center}
\scriptsize

\begin{tabular}{|c|c|c|ccc|ccc|c|ccc|ccc|c|}
\hline
\multirow{3}{*}{\textbf{$\tau$}} & \multirow{3}{*}{\textbf{S}} & \multirow{3}{*}{\textbf{T}} & \multicolumn{7}{c|}{\textbf{Recall}} & \multicolumn{7}{c|}{\textbf{Normalized Cumulative Semantic Recall}} \\ \cline{4-17} 
 &  &  & \multicolumn{3}{c|}{\textbf{I2T}} & \multicolumn{3}{c|}{\textbf{T2I}} & \multicolumn{1}{l|}{} & \multicolumn{3}{c|}{\textbf{I2T}} & \multicolumn{3}{c|}{\textbf{T2I}} & \multicolumn{1}{l|}{} \\ \cline{4-17} 
 &  &  & \textbf{R@1} & \textbf{R@5} & \textbf{R@10} & \textbf{R@1} & \textbf{R@5} & \textbf{R@10} & \textbf{Rsum} & \textbf{N@1} & \textbf{N@5} & \textbf{N@10} & \textbf{N@1} & \textbf{N@5} & \textbf{N@10} & \multicolumn{1}{l|}{\textbf{Nsum}} \\ \hline
3 & SN &\cmark& 76.2 & 93.7 & 96.3 & 63.6 & 91.7 & 96.4 & 517.9 & 69.2 & 73.7 & 69.7 & 75.8 & 72.5 & 68.6 & 429.4 \\
3 & SN &\xmark& 74.8 & 93.5 & 96.8 & 62.2 & 91.5 & 96.2 & 515.0 & 68.3 & 73.2 & 69.4 & 74.8 & 72.7 & 69.0 & 427.3 \\
5 & SN &\cmark& 76.4 & 94.0 & 97.3 & 64.2 & 92.2 & 96.4 & 520.5 & 68.8 & 74.1 & 70.0 & 76.1 & 72.5 & 68.4 & 429.7 \\
5 & SN &\xmark& 76.1 & 93.8 & 96.9 & 63.2 & 91.4 & 96.6 & 518.0 & 68.2 & 73.6 & 69.6 & 75.5 & 72.4 & 68.9 & 428.2 \\
10 & SN &\cmark& 76.9 & 94.2 & 97.7 & 64.4 & 91.8 & 96.3 & 521.3 & 69.9 & 73.8 & 69.6 & 76.3 & 71.9 & 67.7 & 429.1 \\
10 & SN &\xmark& 76.8 & 94.2 & 97.4 & 64.2 & 91.8 & 96.2 & 520.6 & 70.1 & 73.7 & 69.7 & 76.3 & 72.1 & 68.0 & 429.9 \\ \hline
\end{tabular}

\end{center}
\caption{Experiments on MSCOCO 1K regarding the effect of ($\tau$), employing a soft negative sampling strategy (SN). The third column (T) shows whether the original triplet is kept (\cmark) or only our formulation is employed (\xmark). For all our experiments, we employ CVSE\cite{wang2020consensus}. Results are depicted in terms of Recall@K (R@K) and Normalized Cumulative Semantic Score (N@K).
}

\label{tab:coco_ablation1}
\end{table*}

\definecolor{gray}{RGB}{210,210,210}
\newcolumntype{a}{>{\columncolor{gray}}c}

\begin{table*}[!h]
\begin{center}
\scriptsize

\begin{tabular}{|c|c|c|ccc|ccc|c|}
\hline
\multirow{3}{*}{\textbf{$\tau$}} & \multirow{3}{*}{\textbf{S}} & \multirow{3}{*}{\textbf{T}} & \multicolumn{7}{c|}{\textbf{Normalized Cumulative Semantic Recall (Non-GT)}} \\ \cline{4-10} 
 &  &  & \multicolumn{3}{c|}{\textbf{I2T}} & \multicolumn{3}{c|}{\textbf{T2I}} & \multicolumn{1}{l|}{} \\ \cline{4-10} 
 &  &  & \textbf{N@1} & \textbf{N@5} & \textbf{N@10} & \textbf{N@1} & \textbf{N@5} & \textbf{N@10} & \multicolumn{1}{l|}{\textbf{Nsum}} \\ \hline
3 & SN &\cmark& 45.4 & 46.0 & 46.0 & 56.9 & 58.9 & 59.0 & 312.2 \\
3 & SN &\xmark& 45.2 & 46.2 & 46.2 & 56.9 & 59.1 & 59.5 & 313.1 \\
5 & SN &\cmark& 45.4 & 46.2 & 46.2 & 56.6 & 58.6 & 58.6 & 311.5 \\
5 & SN &\xmark& 45.3 & 46.2 & 46.2 & 56.9 & 59.1 & 59.3 & 312.9 \\
10 & SN &\cmark& 44.1 & 45.2 & 45.1 & 56.6 & 57.8 & 57.7 & 306.5 \\
10 & SN &\xmark& 44.4 & 45.7 & 45.5 & 57.1 & 58.3 & 58.2 & 309.3 \\ \hline
\end{tabular}

\end{center}
\caption{Experiments on MSCOCO 1K (Non-GT) regarding the effect of ($\tau$), employing a soft negative sampling strategy (SN). The third column (T) shows whether the original triplet is kept (\cmark) or only our formulation is employed (\xmark). For all our experiments, we employ CVSE\cite{wang2020consensus}. Results are depicted in terms of Recall@K (R@K) and Normalized Cumulative Semantic Score (N@K).}

\label{tab:coco_ablation2}
\end{table*}


\newpage

\section{Qualitative Samples for Reduced Data Scenario}

In this section, we provide qualitative samples on image-to-text and text-to-image in Flickr30K and MSCOCO 1K coming from the reduced data scenario (please refer to Section 5.2 in the main paper) by only using $10$\% of the training set in Flickr30K and $5$\% in MSCOCO. To offer additional insights, we provide the Recall ($R^v$) and NCS per each sample.

\subsection{Image-to-Text}

\begin{table}[htp!]
\centering
\def\arraystretch{1.5}
\begin{tabular}{c|l}

\multirow{6}{*}{\includegraphics[width=0.2\textwidth, height=0.13\textwidth]{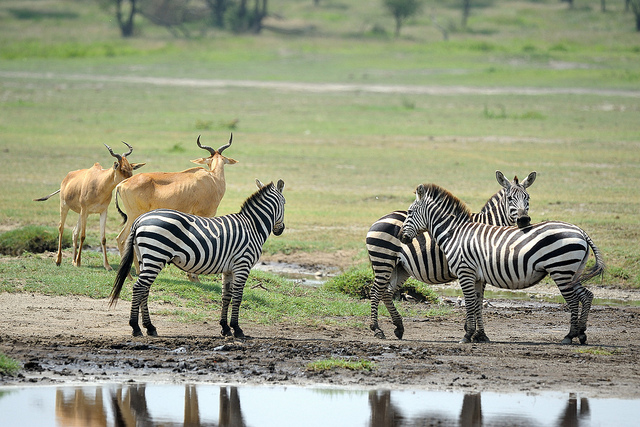}} & \textbf{Ground Truth}:\\ 
& Three zebras and other wild animals out in a semi-green field.
\\
& Three zebras and two other animals grazing. 
\\
& Wildlife standing near water area in natural setting.
\\
& Three zebras near the shore line of a body of water.
\\
& A group of animals stand next to a watering hole.
\\
\hline

\multirow{6}{*}{\textbf{CVSE + SAM}} & \textbf{Three zebras near the shore line of a body of water}. $R_i:1$, $S_i:0.46$
\\
& A heard of zebra on the plains at a watering hole. $R_i:0$, $S_i:0.21$
\\
& \textbf{A group of animals stand next to a watering hole}. $R_i:1$, $S_i:0.33$
\\
& A group of zebras and birds are gathered around water. $R_i:0$, $S_i:0.17$
\\
& There is a herd of zebras standing around. $R_i:0$, $S_i:0.06$
\\
& $R_i@5=1, N_i@5 = 0.61$
\\
\hline

\multirow{6}{*}{\textbf{CVSE}} & \textbf{Three zebras near the shore line of a body of water}. $R_i:1$, $S_i: 0.46$ %
\\
& There is a herd of zebras standing around. $R_i:0$, $S_i:0.06$
\\
& A heard of zebra on the plains at a watering hole. $R_i:0$, $S_i:0.21$
\\
& There are several zebras grazing near the water as a bird flies over them. $R_i:0$, $S_i:0.05$ 
\\
& A group of zebras and birds are gathered around water. $R_i:0$, $S_i:0.17$
\\
& $R_i@5=1, N_i@5 = 0.47$
\\
\hline

\multirow{6}{*}{\textbf{SGR + SAM}} & \textbf{Three zebras near the shore line of a body of water}. $R_i:1$,$S_i:0.46$
\\
& Two zebras fighting in a cloud of dust. $R_i:0$, $S_i:0.05$
\\
& Three zebra in the middle of a field with a body of water in the distance. $R_i:0$, $S_i:0.24$
\\
& Three zebras standing in a sandy desert area. $R_i:0$, $S_i:0.17$
\\
& \textbf{Three zebras and other wild animals out in a semi-green field}. $R_i:1$, $S_i:0.42$
\\
& $R_i@5=1, N_i@5 = 0.67$
\\
\hline
\multirow{6}{*}{\textbf{SGR}} & A zebra grazing on long dry grass in a field. $R_i:0$,$S_i:0.05$
\\
& Four zebras are grazing at a nature reserve. $R_i:0$, $S_i:0.06$
\\
& A group of animals stand next to a watering hole. $R_i:0$, $S_i:0.06$
\\
& Three zebras standing in a sandy desert area. $R_i:0$, $S_i:0.17$
\\
& The small herd of sheep are grazing on the grassy field. $R_i:0$, $S_i:0.05$
\\
& $R_i@5=0, N_i@5 = 0.19$
\\
\hline

\end{tabular}
\caption{Image-to-Text qualitative results in MSCOCO 1K. The initial row depicts the queried image and the associated ground truth captions. Each retrieved caption shows the Recall ($R_i$) and SPICE ($S_i$) score when compared with the GT captions. Each sample showcases the final per sample Recall ($R_i@5$) and NCS ($N_i@5$) score obtained. Bolded captions represent the correctly retrieved ground truth items.}
\label{tab:i2t_coco}
\end{table}

\begin{table}[htp!]
\centering
\begin{tabular}{c|l}

\multirow{6}{*}{\includegraphics[width=0.2\textwidth, height=0.13\textwidth]{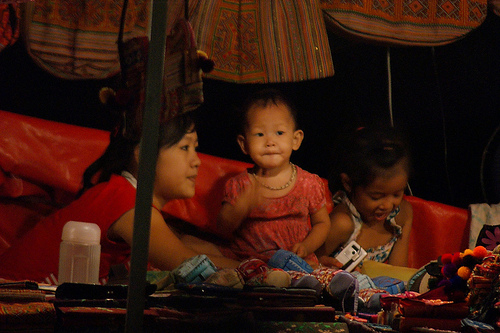}} & \textbf{Ground Truth}:\\ 
& Two children and a woman are sitting on a sofa, one of the children has a camera.
\\
& Three Asian children sitting on a couch with tapestries hanging in the background.
\\
&An Asian woman and her two children sit at a table doing crafts.
\\
& A woman in a red shirt sitting with two young girls in dresses.
\\
& Three young girls in arts and crafts room.
\\
\hline
\multirow{6}{*}{\textbf{CVSE + SAM}} & \textbf{Two children and a woman are sitting on a sofa, one of the children has a camera.} $R_i:1$, $S_i:0.39$
\\
& \textbf{An Asian woman and her two children sit at a table doing crafts.} $R_i:1$, $S_i:0.43$
\\
& \textbf{Three Asian children sitting on a couch with tapestries hanging in the background.} $R_i:1$, $S_i:0.39$
\\
& A woman and three children are in a room full of toys. $R_i:0$, $S_i:0.18$
\\
& A group of children sitting on the floor, eating snacks at school $R_i:0$, $S_i:0.05$
\\
& $R_i@5=1, N_i@5 = 0.72$
\\
\hline

\multirow{6}{*}{\textbf{CVSE}} & Three college-age women sit in upholstered chairs. $R_i:0$, $S_i: 0.05$ %
\\
& Three young women face each other while sitting on red plush chairs. $R_i:0$, $S_i:0.04$
\\
& A plat is sitting on the floor next to a blond girl. $R_i:0$, $S_i:0.05$
\\
& Three girls talking in a lobby. $R_i:0$, $S_i:0.10$ 
\\
& Two kids sitting at a table eating. $R_i:0$, $S_i:0.17$
\\
& $R_i@5=0, N_i@5 = 0.16$
\\
\hline

\multirow{6}{*}{\textbf{SGR + SAM}} & \textbf{Three Asian children sitting on a couch with tapestries hanging in the background.} $R_i:1$,$S_i:0.39$
\\
& \textbf{An Asian woman and her two children sit at a table doing crafts.} $R_i:1$,  $S_i:0.43$
\\
& Six children are sitting around taking notes together. $R_i:0$, $S_i:0.05$
\\
& Woman on four way seesaw with 2 kids. $R_i:0$, $S_i:0.14$
\\
& A group of mostly asian children sitting at cubicles in blue chairs. $R_i:0$, $S_i:0.09$
\\
& $R_i@5=1, N_i@5 = 0.54$ 
\\
\hline

\multirow{6}{*}{\textbf{SGR}} & A child playing in the ocean. $R_i:0$, $S_i:0.05$
\\
& Construction workers deal with removing railroad tracks. $R_i:0$, $S_i:0.00$
\\
& A mural of children on a brick wall. $R_i:0$, $S_i:0.05$
\\
& Four people in the subway are having fun. $R_i:0$, $S_i:0.00$
\\
& Several elderly men are grouped around a table. $R_i:0$, $S_i:0.05$
\\
& $R_i@5=0, N_i@5 = 0.07$
\\
\hline

\end{tabular}
\caption{Image-to-Text qualitative results in Flickr30K. The initial row depicts the queried image and the associated ground truth captions. Each retrieved caption shows the Recall ($R_i$) and SPICE ($S_i$) score when compared with the GT captions. Each sample showcases the final per sample Recall ($R_i@5$) and NCS ($N_i@5$) score obtained. Bold captions represent the correctly retrieved ground truth items. }
\label{tab:i2t_flickr30k}
\end{table}


\newpage
\subsection{Text-to-Image}

\begin{figure*}[h]
    \centering
    \begingroup
    \setlength{\tabcolsep}{2pt} 
    \renewcommand{\arraystretch}{1} 
    \setlength{\fboxsep}{1pt}
    \begin{tabular}{c c c c c}
    \multicolumn{5}{c}{\fontfamily{lmss}\selectfont Query 1: \emph{A group of people on baseball field with trees in the background.}} \\
    \raisebox{1.5\height}[0pt][0pt]{{\rotatebox[origin=c]{90}{CVSE + SAM}}}
    \colorbox{red}{\includegraphics[width=0.17\textwidth, height=0.17\textwidth]{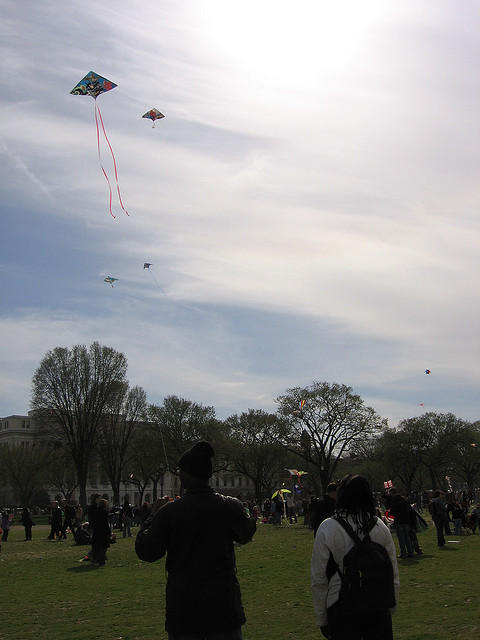}} &
    \colorbox{red}{\includegraphics[width=0.17\textwidth, height=0.17\textwidth]{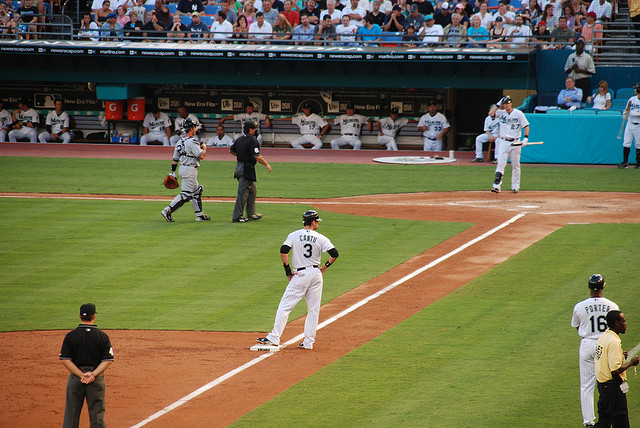}} &
    \colorbox{red}{\includegraphics[width=0.17\textwidth, height=0.17\textwidth]{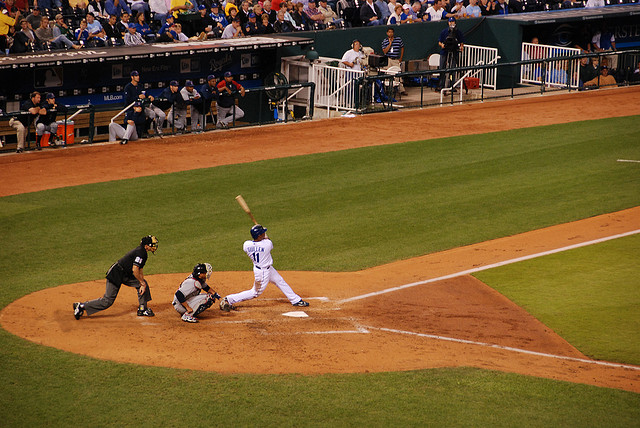}} &
    \colorbox{green}{\includegraphics[width=0.17\textwidth, height=0.17\textwidth]{supp_images/t2i_coco/CVSE_cider/3.jpg}} &
    \colorbox{red}{\includegraphics[width=0.17\textwidth, height=0.17\textwidth]{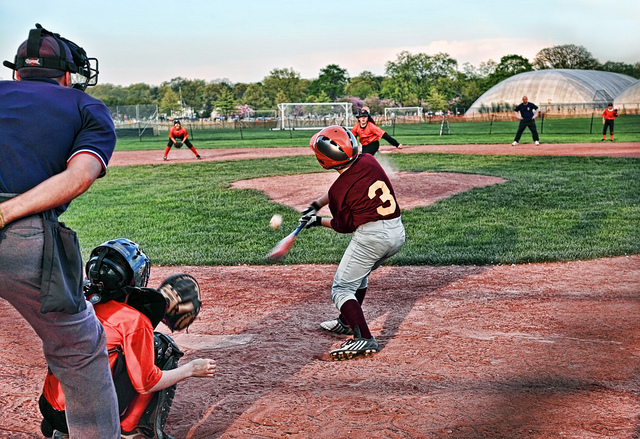}} \\
    $S_i:0.23$ & $S_i:0.07$ & $S_i:0.10$ & $S_i:0.49$ & $S_i:0.06$ \\
    \multicolumn{5}{c}{\fontfamily{lmss}\selectfont $R_i@5=1$, $NCS_i@5 = 0.59$ }
    \\
    
    \multicolumn{5}{c}{}\\ 
    \raisebox{2.5\height}[0pt][0pt]{{\rotatebox[origin=c]{90}{CVSE}}}
    \colorbox{red}{\includegraphics[width=0.17\textwidth, height=0.17\textwidth]{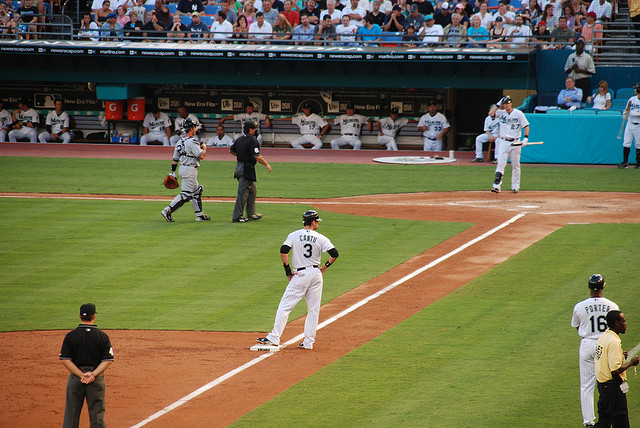}} &
    \colorbox{red}{\includegraphics[width=0.17\textwidth, height=0.17\textwidth]{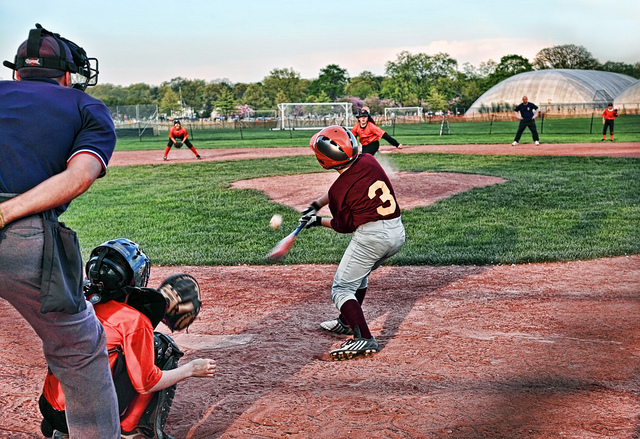}} &
    \colorbox{red}{\includegraphics[width=0.17\textwidth, height=0.17\textwidth]{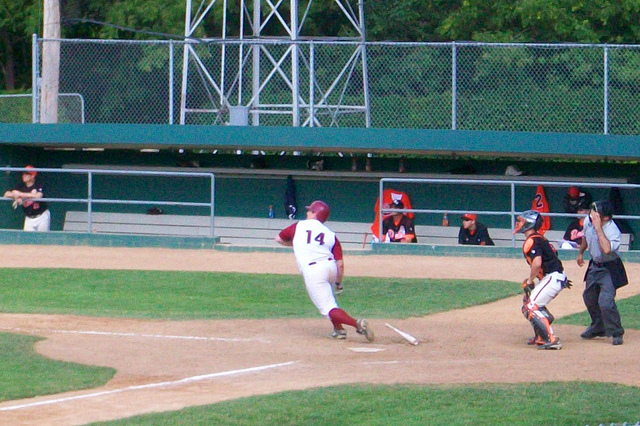}} &
    \colorbox{red}{\includegraphics[width=0.17\textwidth, height=0.17\textwidth]{supp_images/t2i_coco/CVSE_org/3.jpg}} &
    \colorbox{red}{\includegraphics[width=0.17\textwidth, height=0.17\textwidth]{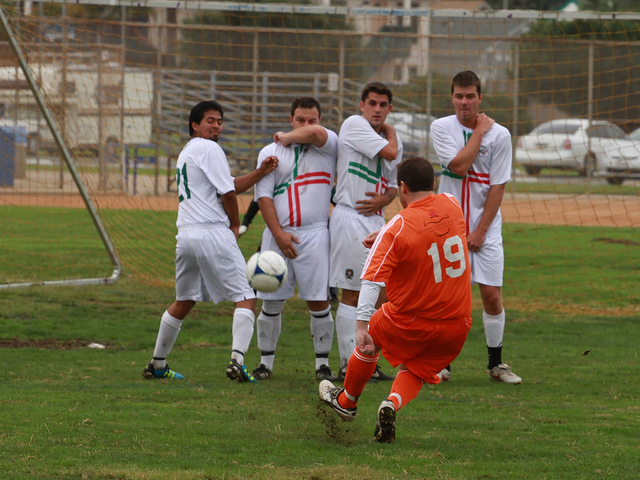}} \\
    $S_i:0.07$ & $S_i:0.06$ & $S_i:0.16$ & $S_i:0.06$ & $S_i:0.25$ \\
    \multicolumn{5}{c}{\fontfamily{lmss}\selectfont $R_i@5=0$, $NCS_i@5 = 0.38$ }\\
    \multicolumn{5}{c}{}\\ 
    \raisebox{1.5\height}[0pt][0pt]{{\rotatebox[origin=c]{90}{SGR + SAM}}}
    \colorbox{red}{\includegraphics[width=0.17\textwidth, height=0.17\textwidth]{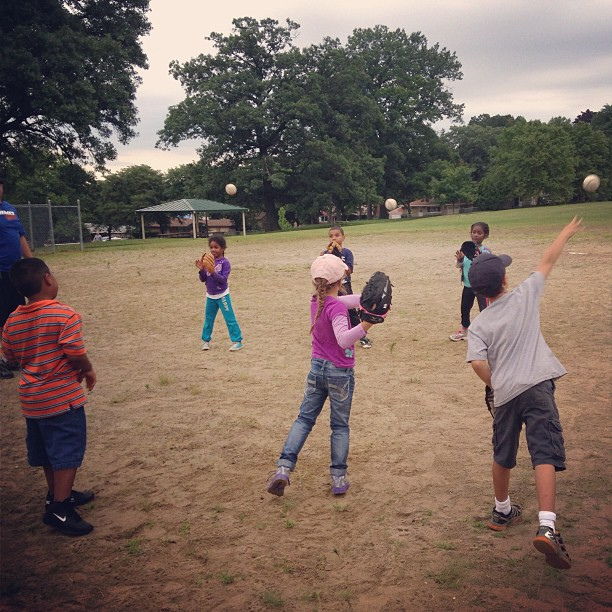}} &
    \colorbox{green}{\includegraphics[width=0.17\textwidth, height=0.17\textwidth]{supp_images/t2i_coco/SGR_cider/1.jpg}} &
    \colorbox{red}{\includegraphics[width=0.17\textwidth, height=0.17\textwidth]{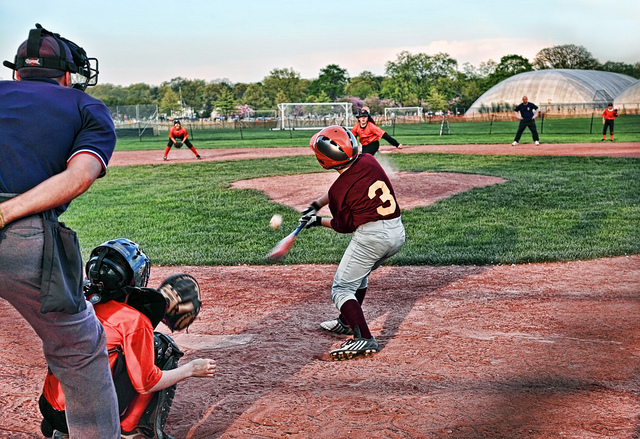}} &
    \colorbox{red}{\includegraphics[width=0.17\textwidth, height=0.17\textwidth]{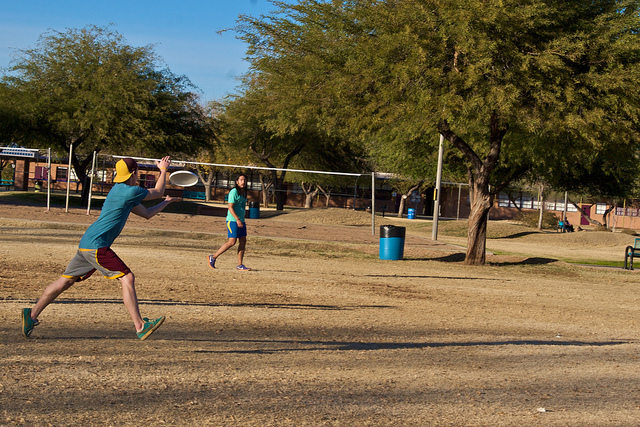}} &
    \colorbox{red}{\includegraphics[width=0.17\textwidth, height=0.17\textwidth]{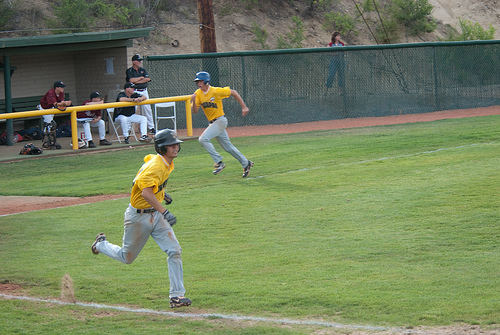}} \\
    $S_i:0.14$ & $S_i:0.49$ & $S_i:0.06$ & $S_i:0.06$ & $S_i:0.11$ \\
    \multicolumn{5}{c}{\fontfamily{lmss}\selectfont $R_i@5=1$, $NCS_i@5 = 0.54$ } \\
    \multicolumn{5}{c}{}\\ 
    \raisebox{3.5\height}[0pt][0pt]{\rotatebox[origin=c]{90}{SGR}}
    \colorbox{red}{\includegraphics[width=0.17\textwidth, height=0.17\textwidth]{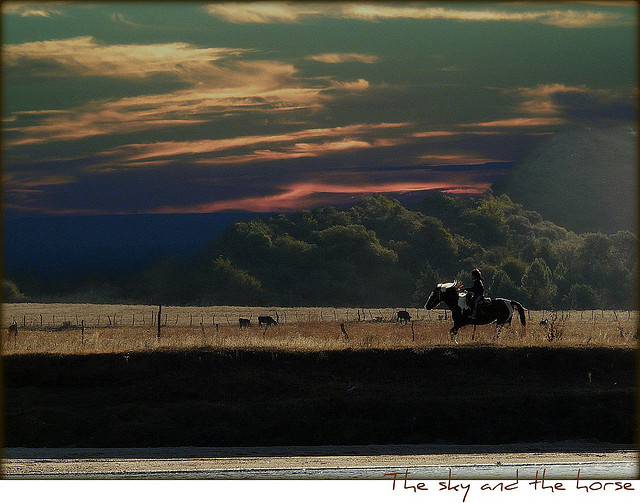}} &
    \colorbox{red}{\includegraphics[width=0.17\textwidth, height=0.17\textwidth]{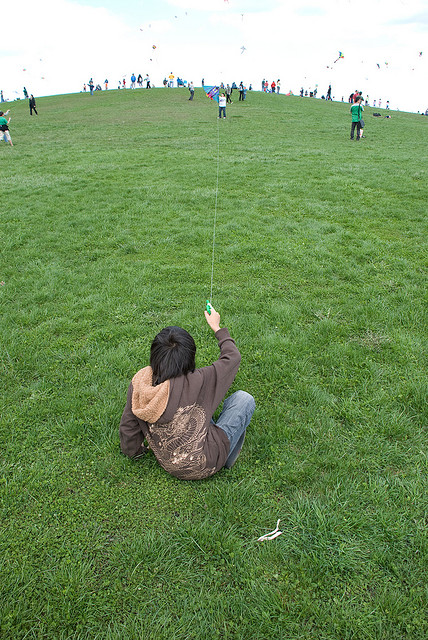}} &
    \colorbox{red}{\includegraphics[width=0.17\textwidth, height=0.17\textwidth]{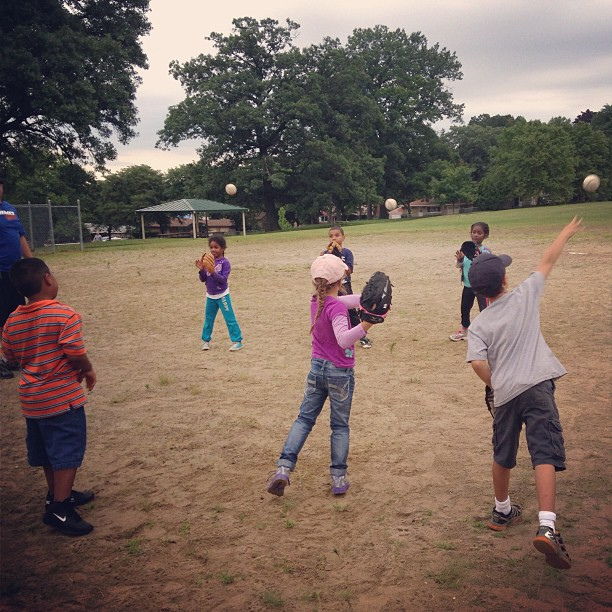}} &
    \colorbox{red}{\includegraphics[width=0.17\textwidth, height=0.17\textwidth]{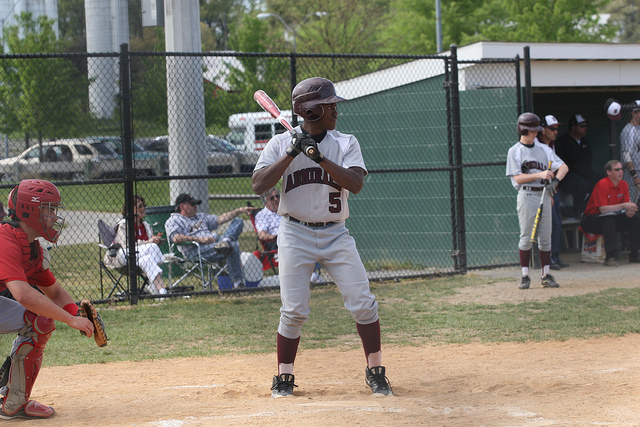}} &
    \colorbox{red}{\includegraphics[width=0.17\textwidth, height=0.17\textwidth]{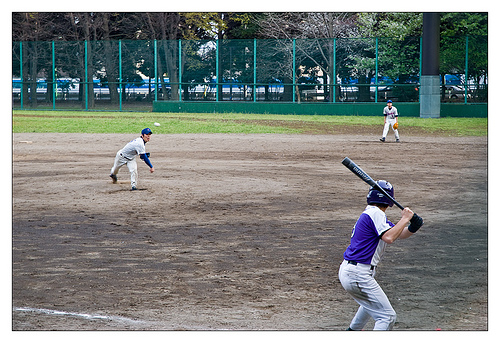}} \\
    $S_i:0.07$ & $S_i:0.31$ & $S_i:0.14$ & $S_i:0.06$ & $S_i:0.00$ \\
    \multicolumn{5}{c}{\fontfamily{lmss}\selectfont $R_i@5=0$, $NCS_i@5 = 0.37$ } \\
    \multicolumn{5}{c}{}\\ 
    \end{tabular}
    \endgroup
    \caption{MSCOCO 1K text-to-image qualitative samples. Each retrieved image shows the SPICE ($S_i$) score when compared with the GT. Recall ($R_i$) is shown as green ($1$) or red ($0$) border on retrieved images. The final score per sample is presented in terms of Recall ($R_i@5$) and NCS ($N_i@5$).}
    \label{fig:qualitative_text2image}
\end{figure*}

\newpage
\begin{figure*}[h]
    \centering
    \begingroup
    \setlength{\tabcolsep}{2pt} 
    \renewcommand{\arraystretch}{1} 
    \setlength{\fboxsep}{1pt}
    \begin{tabular}{c c c c c}
    \multicolumn{5}{c}{\fontfamily{lmss}\selectfont Query 1: \emph{A little girl posing for a picture while eating food.}} \\
    \raisebox{1.5\height}[0pt][0pt]{{\rotatebox[origin=c]{90}{CVSE + SAM}}}
    \colorbox{green}{\includegraphics[width=0.17\textwidth, height=0.17\textwidth]{supp_images2/t2i_coco/CVSE_cider/0.jpg}} &
    \colorbox{red}{\includegraphics[width=0.17\textwidth, height=0.17\textwidth]{supp_images2/t2i_coco/CVSE_cider/1.jpg}} &
    \colorbox{red}{\includegraphics[width=0.17\textwidth, height=0.17\textwidth]{supp_images2/t2i_coco/CVSE_cider/2.jpg}} &
    \colorbox{red}{\includegraphics[width=0.17\textwidth, height=0.17\textwidth]{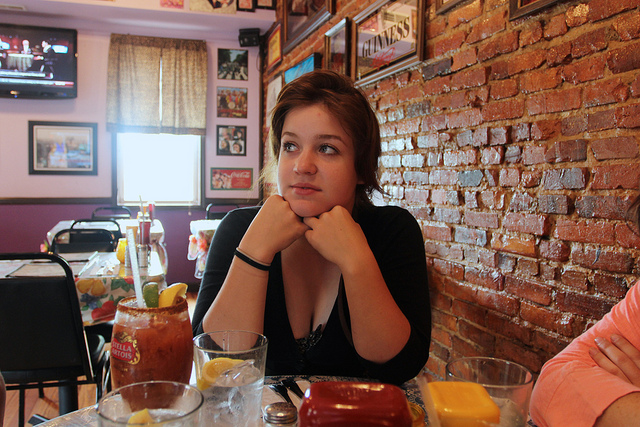}} &
    \colorbox{red}{\includegraphics[width=0.17\textwidth, height=0.17\textwidth]{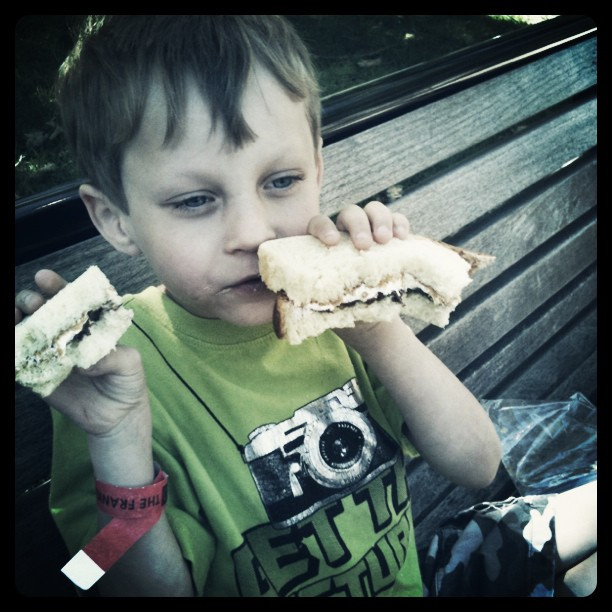}} \\
    $S_i:0.34$ & $S_i:0.14$ & $S_i:0.00$ & $S_i:0.07$ & $S_i:0.00$ \\
    \multicolumn{5}{c}{\fontfamily{lmss}\selectfont $R_i@5=1$, $NCS_i@5 = 0.45$ } \\
    \multicolumn{5}{c}{}\\ 
    \raisebox{2.5\height}[0pt][0pt]{{\rotatebox[origin=c]{90}{CVSE}}}
    \colorbox{green}{\includegraphics[width=0.17\textwidth, height=0.17\textwidth]{supp_images2/t2i_coco/CVSE_org/0.jpg}} &
    \colorbox{red}{\includegraphics[width=0.17\textwidth, height=0.17\textwidth]{supp_images2/t2i_coco/CVSE_org/1.jpg}} &
    \colorbox{red}{\includegraphics[width=0.17\textwidth, height=0.17\textwidth]{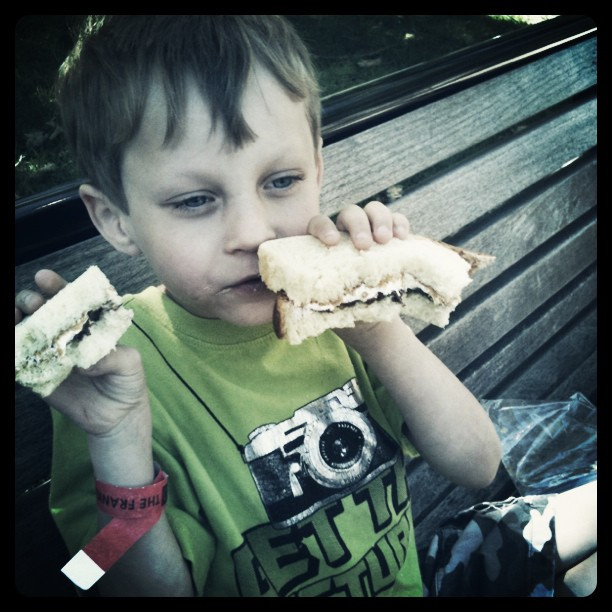}} &
    \colorbox{red}{\includegraphics[width=0.17\textwidth, height=0.17\textwidth]{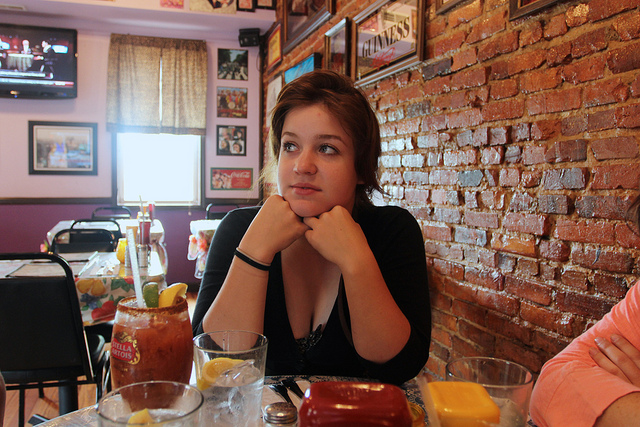}} &
    \colorbox{red}{\includegraphics[width=0.17\textwidth, height=0.17\textwidth]{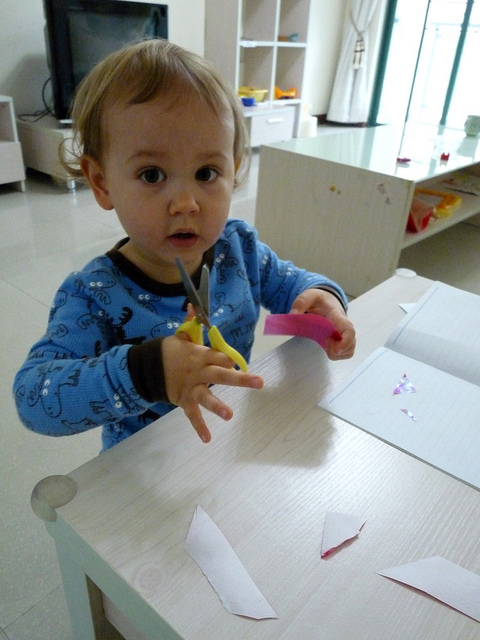}} \\
    $S_i:0.34$ & $S_i:0.00$ & $S_i:0.00$ & $S_i:0.07$ & $S_i:0.00$ \\
    \multicolumn{5}{c}{\fontfamily{lmss}\selectfont $R_i@5=1$, $NCS_i@5 = 0.34$ } \\
    \multicolumn{5}{c}{}\\ 
    \raisebox{1.5\height}[0pt][0pt]{{\rotatebox[origin=c]{90}{SGR + SAM}}}
    \colorbox{green}{\includegraphics[width=0.17\textwidth, height=0.17\textwidth]{supp_images2/t2i_coco/SGR_cider/0.jpg}} &
    \colorbox{red}{\includegraphics[width=0.17\textwidth, height=0.17\textwidth]{supp_images2/t2i_coco/SGR_cider/1.jpg}} &
    \colorbox{red}{\includegraphics[width=0.17\textwidth, height=0.17\textwidth]{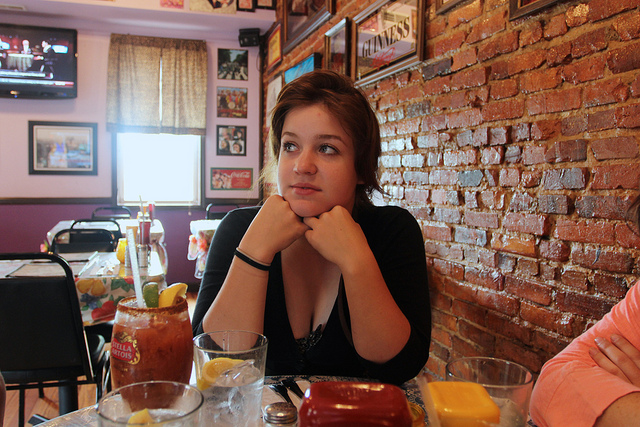}} &
    \colorbox{red}{\includegraphics[width=0.17\textwidth, height=0.17\textwidth]{supp_images2/t2i_coco/SGR_cider/3.jpg}} &
    \colorbox{red}{\includegraphics[width=0.17\textwidth, height=0.17\textwidth]{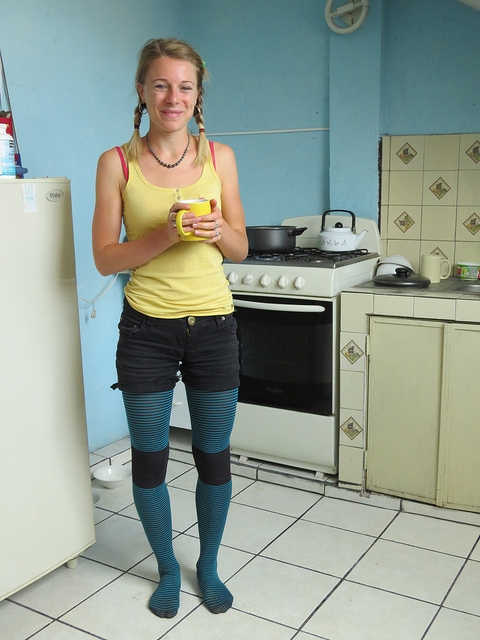}} \\
    $S_i:0.34$ & $S_i:0.14$ & $S_i:0.07$ & $S_i:0.00$ & $S_i:0.11$ \\
    \multicolumn{5}{c}{\fontfamily{lmss}\selectfont $R_i@5=1$, $NCS_i@5 = 0.54$ } \\
    \multicolumn{5}{c}{}\\ 
    \raisebox{3.5\height}[0pt][0pt]{{\rotatebox[origin=c]{90}{SGR}}}
    \colorbox{red}{\includegraphics[width=0.17\textwidth, height=0.17\textwidth]{supp_images2/t2i_coco/SGR_org/0.jpg}} &
    \colorbox{red}{\includegraphics[width=0.17\textwidth, height=0.17\textwidth]{supp_images2/t2i_coco/SGR_org/1.jpg}} &
    \colorbox{red}{\includegraphics[width=0.17\textwidth, height=0.17\textwidth]{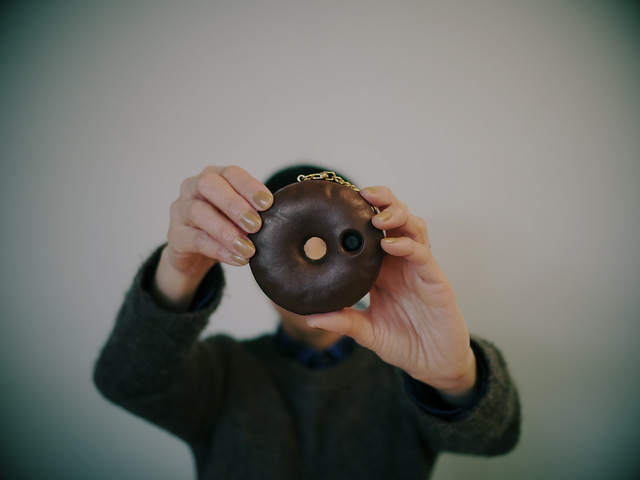}} &
    \colorbox{red}{\includegraphics[width=0.17\textwidth, height=0.17\textwidth]{supp_images2/t2i_coco/SGR_org/3.jpg}} &
    \colorbox{red}{\includegraphics[width=0.17\textwidth, height=0.17\textwidth]{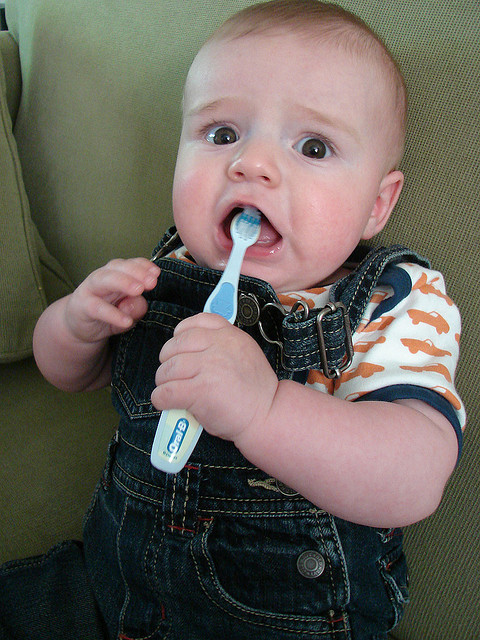}} \\
    $S_i:0.00$ & $S_i:0.14$ & $S_i:0.00$ & $S_i:0.00$ & $S_i:0.00$ \\
    \multicolumn{5}{c}{\fontfamily{lmss}\selectfont $R_i@5=0$, $NCS_i@5 = 0.12$ } \\
    \multicolumn{5}{c}{}\\ 
    \end{tabular}
    \endgroup
    \caption{MSCOCO 1K text-to-image qualitative samples. Each retrieved image shows the SPICE ($S_i$) score when compared with the GT. Recall ($R_i$) is shown as green ($1$) or red ($0$) border on retrieved images. The final score per sample is presented in terms of Recall ($R_i@5$) and NCS ($N_i@5$).}
    \label{fig:qualitative_text2image}
\end{figure*}

\newpage
\begin{figure*}[h]
    \centering
    \begingroup
    \setlength{\tabcolsep}{2pt} 
    \renewcommand{\arraystretch}{1} 
    \setlength{\fboxsep}{1pt}
    \begin{tabular}{c c c c c}
    \multicolumn{5}{c}{\fontfamily{lmss}\selectfont Query 1: \emph{A man is sitting playing guitar.}} \\
    \raisebox{1.5\height}[0pt][0pt]{{\rotatebox[origin=c]{90}{CVSE + SAM}}}
    \colorbox{green}{\includegraphics[width=0.17\textwidth, height=0.17\textwidth]{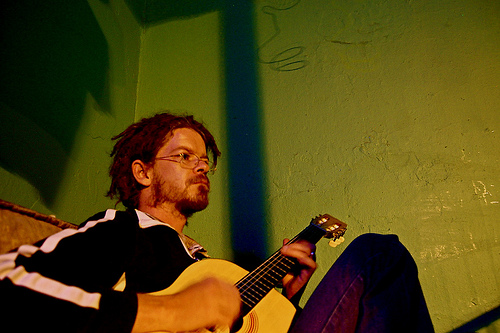}} &
    \colorbox{red}{\includegraphics[width=0.17\textwidth, height=0.17\textwidth]{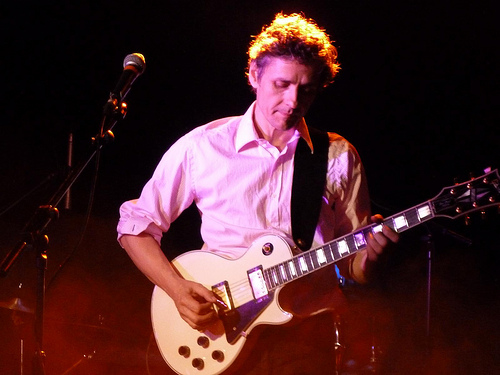}} &
    \colorbox{red}{\includegraphics[width=0.17\textwidth, height=0.17\textwidth]{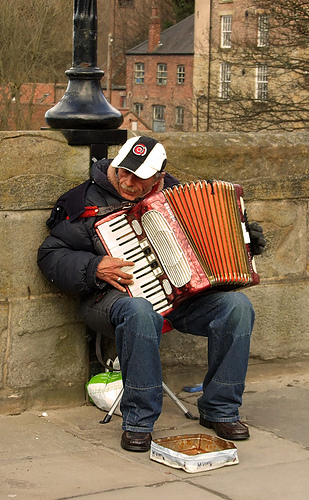}} &
    \colorbox{red}{\includegraphics[width=0.17\textwidth, height=0.17\textwidth]{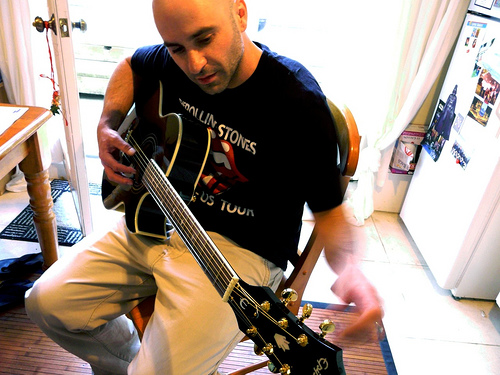}} &
    \colorbox{red}{\includegraphics[width=0.17\textwidth, height=0.17\textwidth]{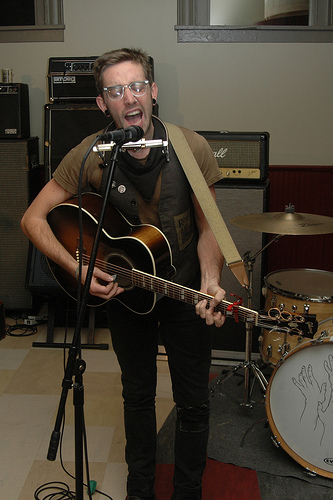}} \\
    $S_i:0.24$ & $S_i:0.09$ & $S_i:0.14$ & $S_i:0.12$ & $S_i:0.24$ \\
    \multicolumn{5}{c}{\fontfamily{lmss}\selectfont $R_i@5=1$, $NCS_i@5 = 0.79$ } \\
    \multicolumn{5}{c}{}\\ 
    \raisebox{2.5\height}[0pt][0pt]{{\rotatebox[origin=c]{90}{CVSE}}}
    \colorbox{red}{\includegraphics[width=0.17\textwidth, height=0.17\textwidth]{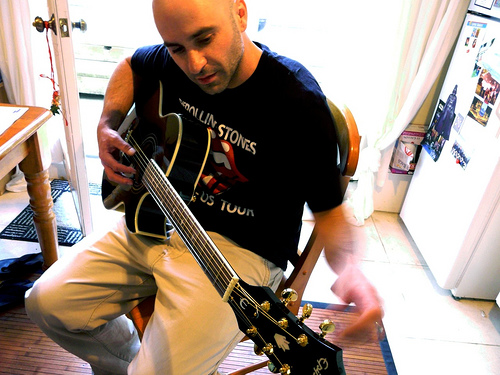}} &
    \colorbox{red}{\includegraphics[width=0.17\textwidth, height=0.17\textwidth]{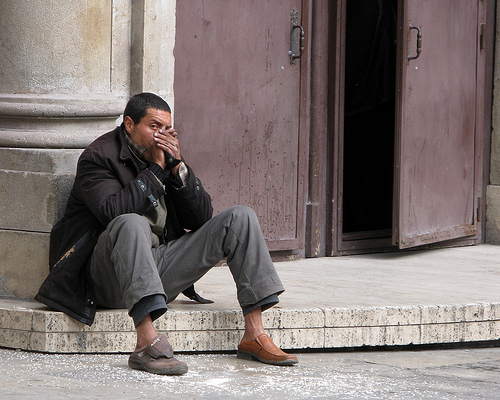}} &
    \colorbox{green}{\includegraphics[width=0.17\textwidth, height=0.17\textwidth]{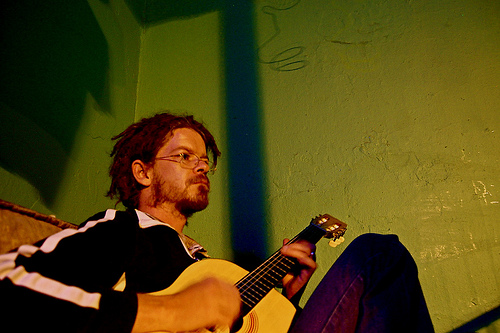}} &
    \colorbox{red}{\includegraphics[width=0.17\textwidth, height=0.17\textwidth]{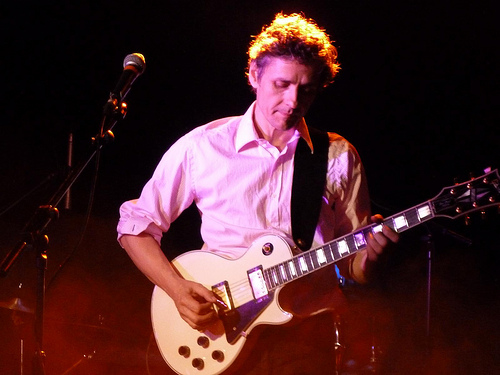}} &
    \colorbox{red}{\includegraphics[width=0.17\textwidth, height=0.17\textwidth]{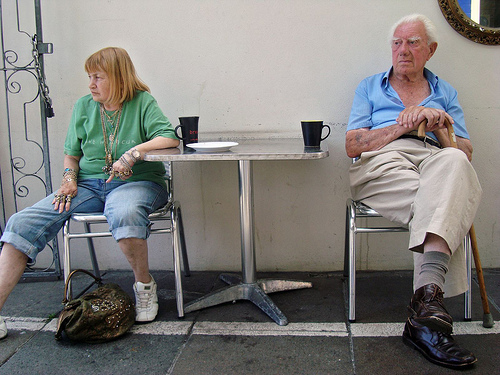}} \\
    $S_i:0.12$ & $S_i:0.06$ & $S_i:0.24$ & $S_i:0.09$ & $S_i:0.07$ \\
    \multicolumn{5}{c}{\fontfamily{lmss}\selectfont $R_i@5=1$, $NCS_i@5 = 0.56$ } \\
    \multicolumn{5}{c}{}\\ 
    \raisebox{1.5\height}[0pt][0pt]{{\rotatebox[origin=c]{90}{SGR + SAM}}}
    \colorbox{green}{\includegraphics[width=0.17\textwidth, height=0.17\textwidth]{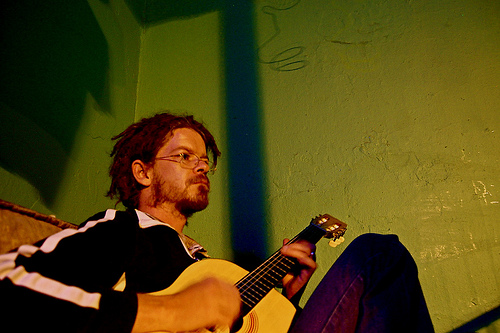}} &
    \colorbox{red}{\includegraphics[width=0.17\textwidth, height=0.17\textwidth]{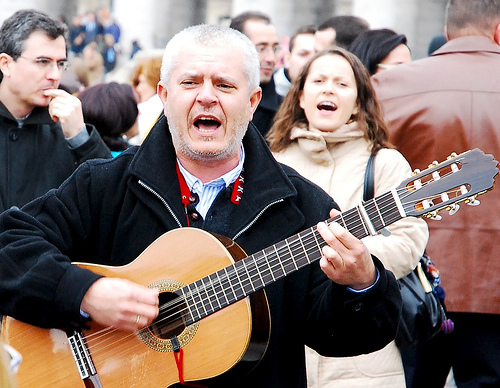}} &
    \colorbox{red}{\includegraphics[width=0.17\textwidth, height=0.17\textwidth]{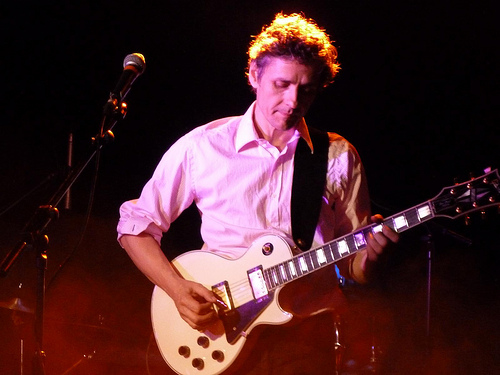}} &
    \colorbox{red}{\includegraphics[width=0.17\textwidth, height=0.17\textwidth]{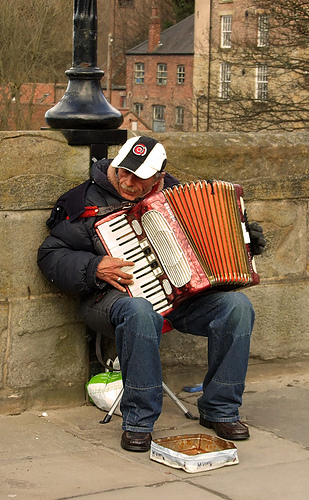}} &
    \colorbox{red}{\includegraphics[width=0.17\textwidth, height=0.17\textwidth]{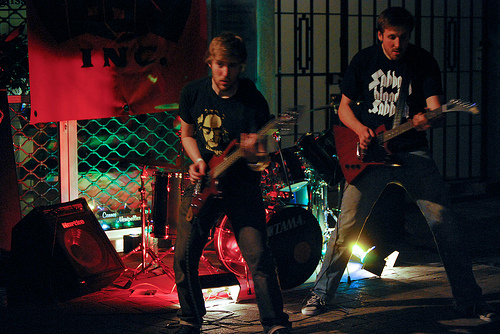}} \\
    $S_i:0.24$ & $S_i:0.18$ & $S_i:0.09$ & $S_i:0.14$ & $S_i:0.19$ \\
    \multicolumn{5}{c}{\fontfamily{lmss}\selectfont $R_i@5=1$, $NCS_i@5 = 0.81$ } \\
    \multicolumn{5}{c}{}\\ 
    \raisebox{3.5\height}[0pt][0pt]{{\rotatebox[origin=c]{90}{SGR}}}
    \colorbox{red}{\includegraphics[width=0.17\textwidth, height=0.17\textwidth]{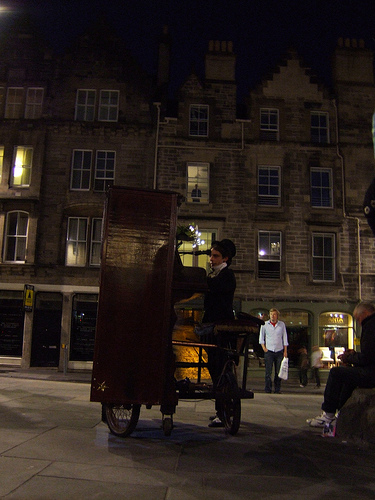}} &
    \colorbox{red}{\includegraphics[width=0.17\textwidth, height=0.17\textwidth]{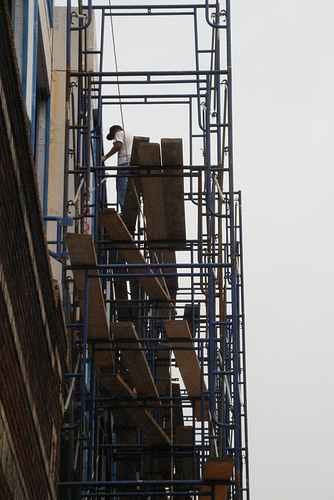}} &
    \colorbox{red}{\includegraphics[width=0.17\textwidth, height=0.17\textwidth]{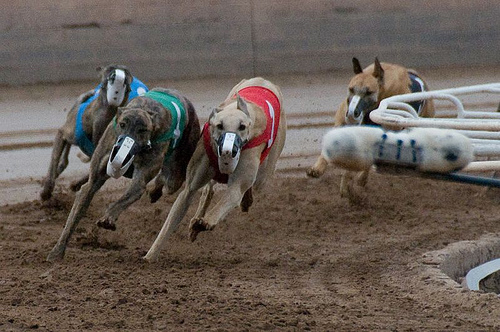}} &
    \colorbox{red}{\includegraphics[width=0.17\textwidth, height=0.17\textwidth]{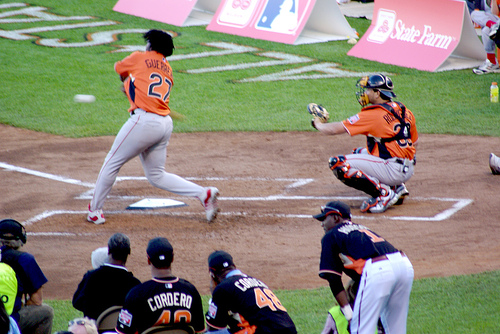}} &
    \colorbox{red}{\includegraphics[width=0.17\textwidth, height=0.17\textwidth]{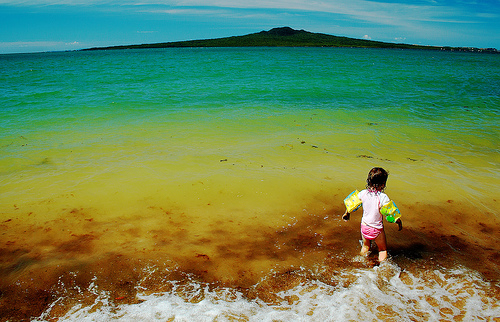}} \\
    $S_i:0.06$ & $S_i:0.06$ & $S_i:0.00$ & $S_i:0.06$ & $S_i:0.00$ \\
    \multicolumn{5}{c}{\fontfamily{lmss}\selectfont $R_i@5=0$, $NCS_i@5 = 0.18$ } \\
    \multicolumn{5}{c}{}\\ 
    \end{tabular}
    \endgroup
    \caption{Flickr30K text-to-image qualitative samples. Each retrieved image shows the SPICE ($S_i$) score when compared with the GT. Recall ($R_i$) is shown as green ($1$) or red ($0$) border on retrieved images. The final score per sample is presented in terms of Recall ($R_i@5$) and NCS ($N_i@5$).}
    \label{fig:qualitative_text2image}
\end{figure*}

\newpage
\begin{figure*}[h]
    \centering
    \begingroup
    \setlength{\tabcolsep}{2pt} 
    \renewcommand{\arraystretch}{1} 
    \setlength{\fboxsep}{1pt}
    \begin{tabular}{c c c c c}
    \multicolumn{5}{c}{\fontfamily{lmss}\selectfont Query 1: \emph{People walking on a trail in a tree filled park.}} \\
    \raisebox{1.5\height}[0pt][0pt]{{\rotatebox[origin=c]{90}{CVSE + SAM}}}
    \colorbox{red}{\includegraphics[width=0.17\textwidth, height=0.17\textwidth]{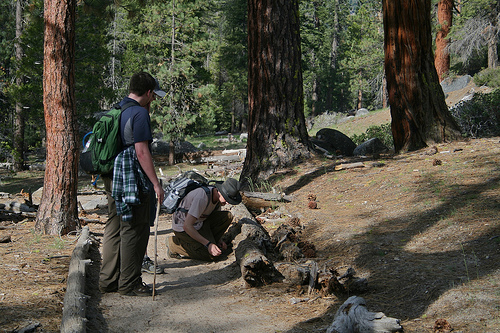}} &
    \colorbox{red}{\includegraphics[width=0.17\textwidth, height=0.17\textwidth]{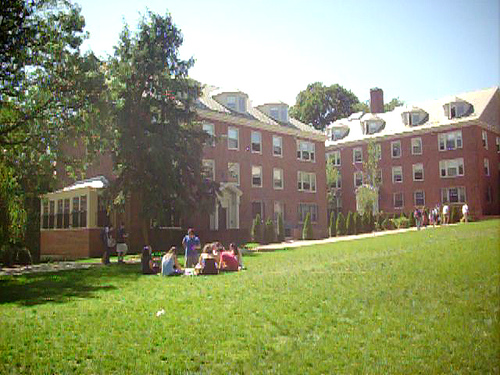}} &
    \colorbox{green}{\includegraphics[width=0.17\textwidth, height=0.17\textwidth]{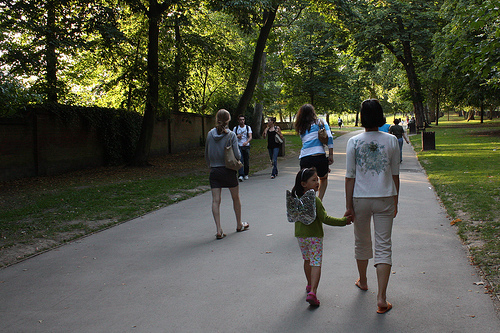}} &
    \colorbox{red}{\includegraphics[width=0.17\textwidth, height=0.17\textwidth]{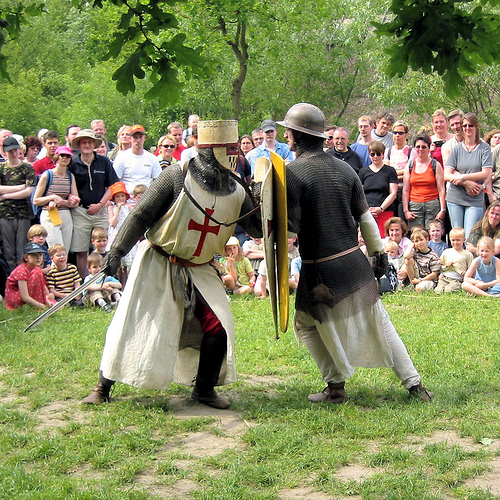}} &
    \colorbox{red}{\includegraphics[width=0.17\textwidth, height=0.17\textwidth]{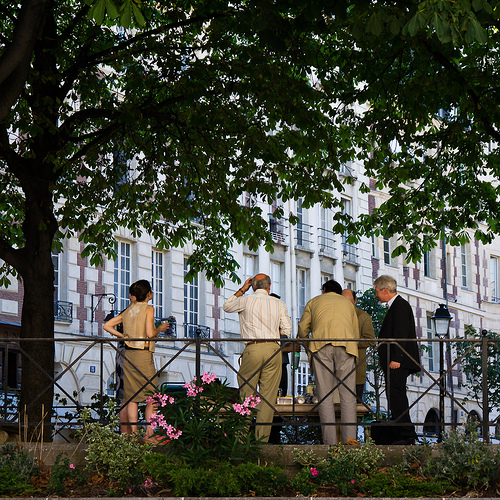}} \\
    $S_i:0.00$ & $S_i:0.06$ & $S_i:0.31$ & $S_i:0.06$ & $S_i:0.11$ \\
    \multicolumn{5}{c}{\fontfamily{lmss}\selectfont $R_i@5=1$, $NCS_i@5 = 0.52$ } \\
    \multicolumn{5}{c}{}\\ 
    \raisebox{2.5\height}[0pt][0pt]{{\rotatebox[origin=c]{90}{CVSE}}}
    \colorbox{red}{\includegraphics[width=0.17\textwidth, height=0.17\textwidth]{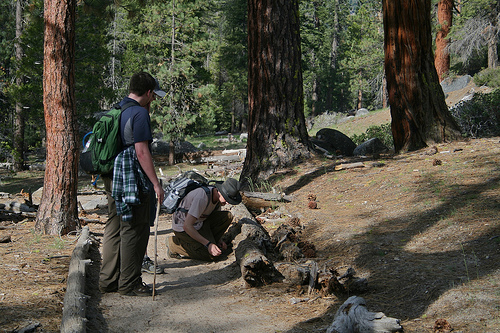}} &
    \colorbox{red}{\includegraphics[width=0.17\textwidth, height=0.17\textwidth]{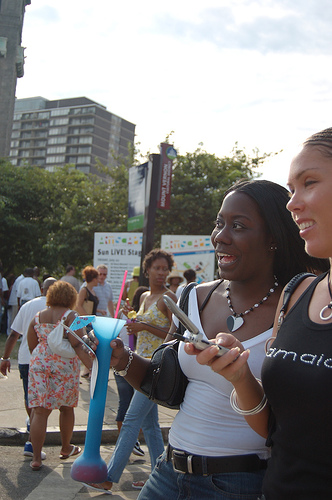}} &
    \colorbox{red}{\includegraphics[width=0.17\textwidth, height=0.17\textwidth]{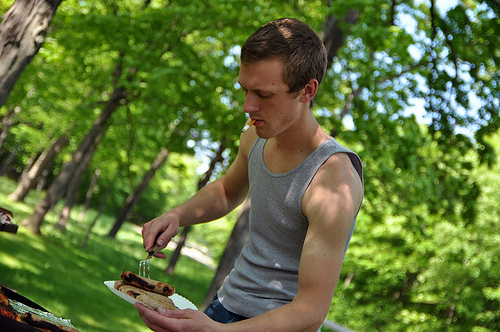}} &
    \colorbox{red}{\includegraphics[width=0.17\textwidth, height=0.17\textwidth]{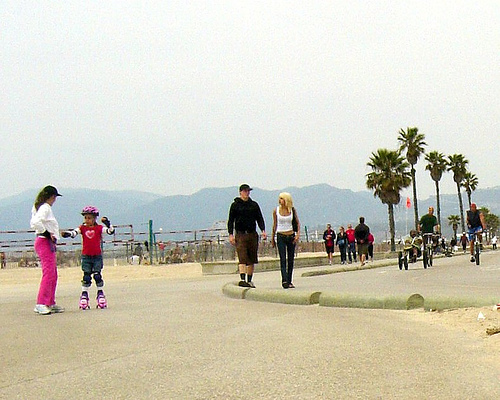}} &
    \colorbox{red}{\includegraphics[width=0.17\textwidth, height=0.17\textwidth]{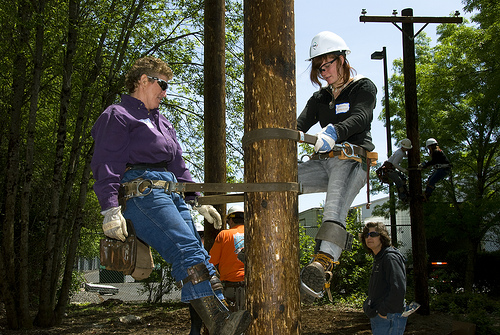}} \\
    $S_i:0.00$ & $S_i:0.04$ & $S_i:0.00$ & $S_i:0.14$ & $S_i:0.04$ \\
    \multicolumn{5}{c}{\fontfamily{lmss}\selectfont $R_i@5=0$, $NCS_i@5 = 0.21$ } \\
    \multicolumn{5}{c}{}\\ 
    \raisebox{1.5\height}[0pt][0pt]{{\rotatebox[origin=c]{90}{SGR + SAM}}}
    \colorbox{red}{\includegraphics[width=0.17\textwidth, height=0.17\textwidth]{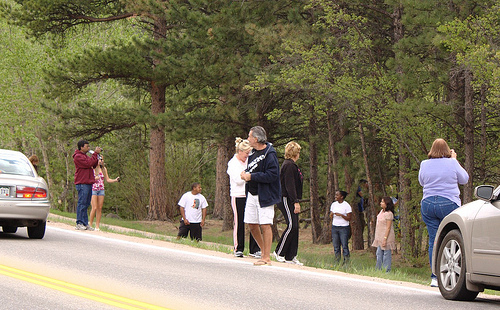}} &
    \colorbox{green}{\includegraphics[width=0.17\textwidth, height=0.17\textwidth]{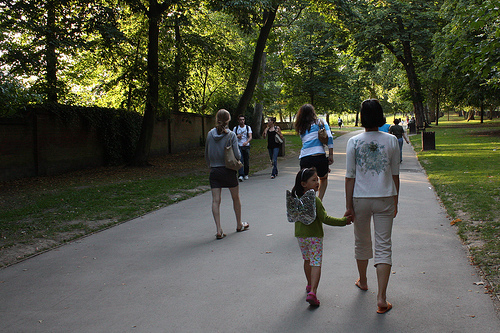}} &
    \colorbox{red}{\includegraphics[width=0.17\textwidth, height=0.17\textwidth]{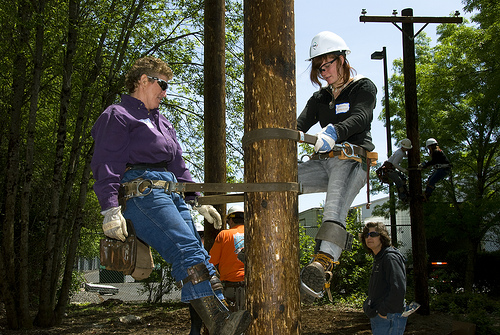}} &
    \colorbox{red}{\includegraphics[width=0.17\textwidth, height=0.17\textwidth]{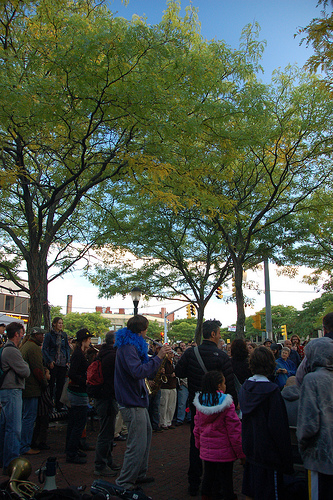}} &
    \colorbox{red}{\includegraphics[width=0.17\textwidth, height=0.17\textwidth]{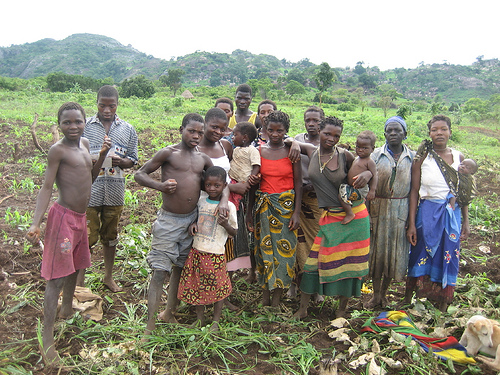}} \\
    $S_i:0.06$ & $S_i:0.31$ & $S_i:0.04$ & $S_i:0.17$ & $S_i:0.12$ \\
    \multicolumn{5}{c}{\fontfamily{lmss}\selectfont $R_i@5=1$, $NCS_i@5 = 0.66$ } \\
    \multicolumn{5}{c}{}\\ 
    \raisebox{3.5\height}[0pt][0pt]{{\rotatebox[origin=c]{90}{SGR}}}
    \colorbox{red}{\includegraphics[width=0.17\textwidth, height=0.17\textwidth]{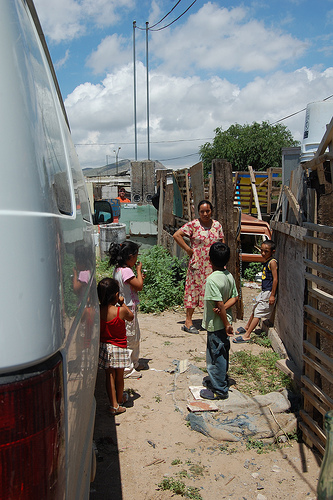}} &
    \colorbox{red}{\includegraphics[width=0.17\textwidth, height=0.17\textwidth]{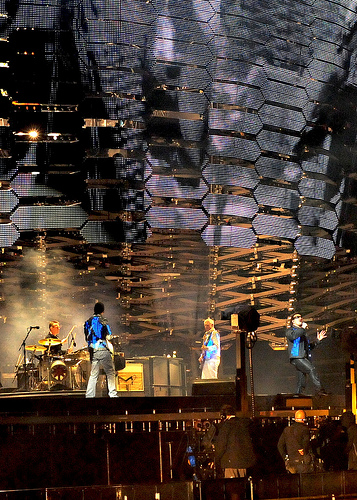}} &
    \colorbox{red}{\includegraphics[width=0.17\textwidth, height=0.17\textwidth]{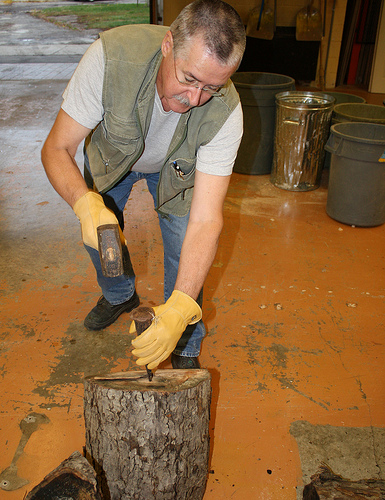}} &
    \colorbox{red}{\includegraphics[width=0.17\textwidth, height=0.17\textwidth]{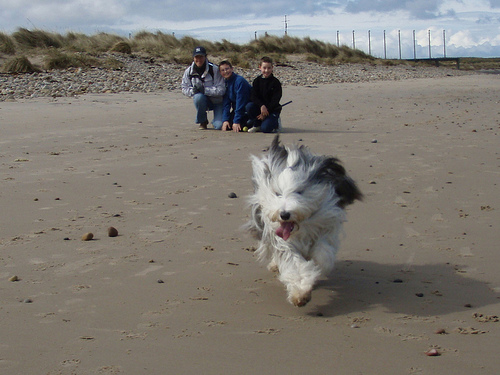}} &
    \colorbox{red}{\includegraphics[width=0.17\textwidth, height=0.17\textwidth]{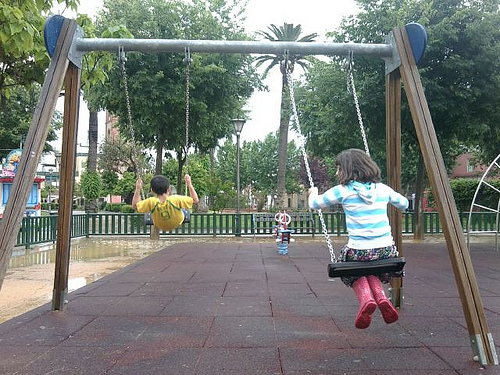}} \\
    $S_i:0.00$ & $S_i:0.00$ & $S_i:0.06$ & $S_i:0.12$ & $S_i:0.11$ \\
    \multicolumn{5}{c}{\fontfamily{lmss}\selectfont $R_i@5=0$, $NCS_i@5 = 0.27$ } \\
    \multicolumn{5}{c}{}\\ 
    \end{tabular}
    \endgroup
    \caption{Flickr30K text-to-image qualitative samples. Each retrieved image shows the SPICE ($S_i$) score when compared with the GT. Recall ($R_i$) is shown as green ($1$) or red ($0$) border on retrieved images. The final score per sample is presented in terms of Recall ($R_i@5$) and NCS ($N_i@5$).}
    \label{fig:qualitative_text2image}
\end{figure*}
\clearpage
\newpage
\section{Insights on State-of-the-Art Retrieval}
In this section, we extend the performance results of state-of-the-art ITM retrieval models. Similarly as in the main paper, the following models are evaluated: VSE++~\cite{faghri2017vse++}, SCAN~\cite{lee2018stacked}, VSRN~\cite{li2019visual}, CVSE~\cite{wang2020consensus}, SGR and SAF~\cite{diao2021similarity}. The figures comparing the models at different cut-off points (@k), evaluated with and without GT samples are shown in Table~\ref{table:full_reality_check}. 
The same behaviour described in the main document can be observed. As the number of relevant images increase, the Recall(R) depicts a constant score increase, while the Semantic Recall (SR) shows the opposite. 

On one hand, state-of-the-art models tend to miss semantic relevant samples that are not labelled in the dataset as ground truth. 
On another hand, even though there is a significant performance gap between recently proposed (CVSE, SAF, SGR) and previous (VSE++, SCAN) models when they are evaluated in terms of Recall, this statement does not necessarily hold. The aforementioned effect is even more significant when models are evaluated on the non-GT annotations.
It is important to mention that aside missing $4$ annotated GT samples in the image-to-text scenario, the limitations of evaluating with Recall $R^v$ are present at the moment of considering relevant samples. As an overview, these results suggest that there is still a considerable room for improvement in ITM pipelines.

\afterpage{%
\clearpage
    \thispagestyle{empty}
    \atxy{\dimexpr\paperwidth-1in}{.5\paperheight}{\rotatebox[origin=center]{90}{\thepage}}
\begin{landscape}
\begin{table}[htp!]
\centering
\resizebox{1.4\textwidth}{!}{
\begin{tabular}{cc|c|c}
         & {\fontsize{50}{0}\selectfont SR@1 \& R@1} & \fontsize{50}{0}\selectfont{SR@5 \& R@5} & \fontsize{50}{0}\selectfont{SR@10 \& R@10}\\
      \raisebox{3.5\height}[0pt][0pt]{\rotatebox[origin=c]{90}{\fontsize{50}{0}\selectfont Image2Text}} &
      \includegraphics[height=1.2\linewidth,width=1.3\linewidth]{supp_figures/reality_check_@1_i2t_non_gt.pdf} &   \includegraphics[height=1.2\linewidth,width=1.3\linewidth]{supp_figures/reality_check_@5_i2t_non_gt.pdf} &  \includegraphics[height=1.2\linewidth,width=1.3\linewidth]{supp_figures/reality_check_@10_i2t_non_gt.pdf} \\
    \raisebox{3.5\height}[0pt][0pt]{\rotatebox[origin=c]{90}{\fontsize{50}{0}\selectfont Text2Image}} & \includegraphics[height=1.2\linewidth,width=1.3\linewidth]{supp_figures/reality_check_@1_t2i_non_gt.pdf} &   \includegraphics[height=1.2\linewidth,width=1.3\linewidth]{supp_figures/reality_check_@5_t2i_non_gt.pdf} &  \includegraphics[height=1.2\linewidth,width=1.3\linewidth]{supp_figures/reality_check_@10_t2i_non_gt.pdf} \\
\end{tabular}
}

\captionsetup{width=0.9\linewidth}
\caption{Results evaluated with Recall(R) and the the proposed Semantic Recall(SR) for GT and Non-GT items. Each column represents the figures obtained at a cut-off point @1, @5 and @10 respectively. Best viewed in color.}
\label{table:full_reality_check}
\end{table}
\end{landscape}

\clearpage
\thispagestyle{empty}
\atxy{\dimexpr\paperwidth-1in}{.5\paperheight}{\rotatebox[origin=center]{90}{\thepage}}
\begin{landscape}
\begin{table}[!htp]
\centering
\resizebox{1.4\textwidth}{!}{
\begin{tabular}{cc|c|c}
     & {\fontsize{50}{0}\selectfont SR@1 \& R@1} & \fontsize{50}{0}\selectfont{SR@5 \& R@5} & \fontsize{50}{0}\selectfont{SR@10 \& R@10}\\
     \raisebox{3.5\height}[0pt][0pt]{\rotatebox[origin=c]{90}{\fontsize{50}{0}\selectfont Image2Text}} &\includegraphics[height=1.2\linewidth,width=1.3\linewidth]{supp_figures/reality_check_@1_i2t_gt.pdf} &   \includegraphics[height=1.2\linewidth,width=1.3\linewidth]{supp_figures/reality_check_@5_i2t_gt.pdf} &  \includegraphics[height=1.2\linewidth,width=1.3\linewidth]{supp_figures/reality_check_@10_i2t_gt.pdf} \\
    \raisebox{3.5\height}[0pt][0pt]{\rotatebox[origin=c]{90}{\fontsize{50}{0}\selectfont Text2Image}} & \includegraphics[height=1.2\linewidth,width=1.3\linewidth]{supp_figures/reality_check_@1_t2i_gt.pdf} &   \includegraphics[height=1.2\linewidth,width=1.3\linewidth]{supp_figures/reality_check_@5_t2i_gt.pdf} &  \includegraphics[height=1.2\linewidth,width=1.3\linewidth]{supp_figures/reality_check_@10_t2i_gt.pdf}
\end{tabular}
}
\captionsetup{width=0.9\linewidth}
\caption{Results evaluated with Recall(R) and the the proposed Semantic Recall(SR) for GT and Non-GT items. Each column represents the figures obtained at a cut-off point @1, @5 and @10 respectively. Best viewed in color.}

\label{table:full_reality_check_2}
\end{table}
\end{landscape}
}

\newpage

\clearpage
{\small
\bibliographystyle{ieee_fullname}
\bibliography{egbib}
}